%% file: main.tex
\documentclass[twocolumn,letterpaper]{IEEEAerospaceCLS}  

\usepackage[pdftex,final]{graphicx}    
\newcommand{\ignore}[1]{}  

\usepackage{caption}
\usepackage{subcaption}
\usepackage{amsmath}
\usepackage{tikz}
\usepackage{dsfont}

\usepackage{xcolor}
\usepackage[pagebackref=true,colorlinks]{hyperref}
\renewcommand*{\backref}[1]{%
  \ifx#1\relax
  \else
    (Page #1)%
  \fi
}
\hypersetup{
  colorlinks   = true, 
  urlcolor     = red, 
  linkcolor    = blue, 
  citecolor   = red 
}

\begin{document}

\title{Icy Moon Surface Simulation and Stereo Depth Estimation for Sampling Autonomy}

\author{%
Ramchander Bhaskara\textsuperscript{1}, Georgios Georgakis\textsuperscript{2}, Jeremy Nash\textsuperscript{2}, Marissa Cameron\textsuperscript{2}, Joseph Bowkett\textsuperscript{2}, 
\\ Adnan Ansar\textsuperscript{2}, Manoranjan Majji \textsuperscript{1}, Paul Backes\textsuperscript{2}\\
\\
\textsuperscript{1}Dept. of Aerospace Engineering, Texas A\&M University, College Station, TX 77843 \\
bhaskara@tamu.edu
\and
\textsuperscript{2}Jet Propulsion Laboratory, California Institute of Technology, Pasadena, CA 91109\\
georgios.georgakis@jpl.nasa.gov
\\
\thanks{\footnotesize 979-8-3503-0462-6/24/$\$31.00$ \copyright2024 IEEE} 
}

\maketitle

\thispagestyle{plain}
\pagestyle{plain}

\newpage

\begin{abstract}

Sampling autonomy for icy moon lander missions requires understanding of topographic and photometric properties of the sampling terrain. Unavailability of high resolution visual datasets (either bird-eye view or point-of-view from a lander) is an obstacle for selection, verification or development of perception systems. We attempt to alleviate this problem by: 1) proposing Graphical Utility for Icy moon Surface Simulations (GUISS) framework, for versatile stereo dataset generation that spans the spectrum of bulk photometric properties, and 2) focusing on a stereo-based visual perception system and evaluating both traditional and deep learning-based algorithms for depth estimation from stereo matching. 
The surface reflectance properties of icy moon terrains (Enceladus and Europa) are inferred from multispectral datasets of previous missions. With procedural terrain generation and physically valid illumination sources, our framework can fit a wide range of hypotheses with respect to visual representations of icy moon terrains. This is followed by a study over the performance of stereo matching algorithms under different visual hypotheses. Finally, we emphasize the standing challenges to be addressed for simulating perception data assets for icy moons such as Enceladus and Europa. Our code can be found here:  \url{https://github.com/nasa-jpl/guiss}.

\end{abstract}

\tableofcontents

\input{Sections/01_intro}
\input{Sections/02_DesignGoals}
\input{Sections/03_RelatedWork}
\input{Sections/04_pipeline}

\input{Sections/05_stereo}
\input{Sections/06_Conclusions}

\input{Sections/07_FutureWork}
\input{Sections/08_Acknowledgements}

\bibliographystyle{IEEEtran}
\bibliography{Sections/References} 

\input{Sections/biography}
\end{document}

%% file: Sections/01_intro.tex
\section{Introduction}




\begin{figure}
\centering
\includegraphics[width=3in]{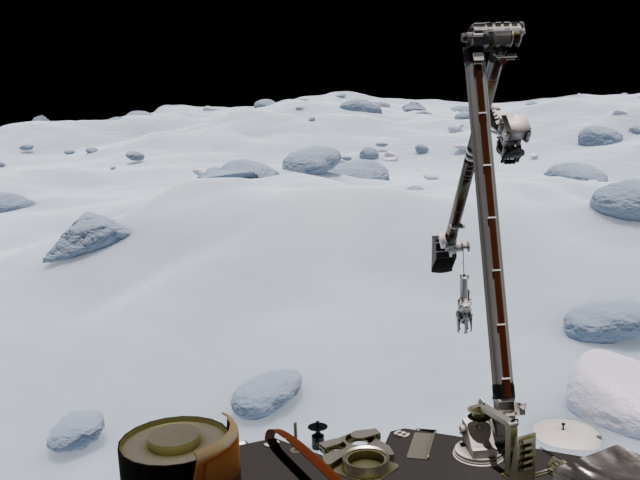} 
\caption{Representative simulation of icy moon terrain rendered using {GUISS}. }
\label{Fig:landerVisualization}
\end{figure}

Robotic exploration of ocean worlds such as Europa and Enceladus, 
has astrobiological, geochemical, and geophysical significance for detecting conditions conducive to extraterrestrial life. Evidence indicating hydrothermal activities on Europa's and Enceladus' oceanic subsurface are corroborated through plume eruption observations from Hubble Space Telescope \cite{roth2014transient} and Cassini flyby \cite{hansen2006enceladus} respectively among other measurements. The 2023-2032 Planetary Science Decadal Survey proposes Enceladus Orbilander as the second highest priority new Flagship mission \cite{NAP26522}. 
The Europa Lander mission is a strategic concept with science objectives that include investigation of the potential for life in the subsurface ocean below Europa's ice crust \cite{dooley2018mission,hand2017europa}.
Jupiter Icy Moons Explorer (JUICE) \cite{grasset2013jupiter} and Europa Clipper \cite{bayer2019europa} missions are currently being realized to provide a foundation for future in-situ ocean world exploration.




Ocean world lander missions are quintessentially tasked with objectives that include monitoring the environment as well as collecting in-situ samples in search of habitability information and biosignatures. 
The sampling activities for the icy moons include robotically excavating material from or beneath the surface depending upon the sampling environment and corresponding science objectives. Sampling at depths of $10-30$ cms is of interest for Europan environment \cite{nordheim2018preservation}. 
Collecting micron-sized ice grains from the plumes is of interest on Enceladus \cite{postberg2011salt,buratti2022observations}. The likely salt-rich granular water-ice particles form a compositional stratification of the plume with higher proportion of fresh organic material closer to the surface \cite{postberg2011salt}.  
While harsh radiated and geothermal environments limit the mission life, constraints around communication latency and limited power from batteries pose significant operational challenges for sampling. Therefore, tight coupling between perception and sampling autonomy is paramount to meet the science objectives \cite{nesnas2021autonomy}. 

Several developments for the icy-moon lander sampling technology are being carried out at Jet Propulsion Lab (JPL) to produce and test robotic sampling mechanisms. Ocean World Lander Autonomy Testbed (OWLAT) is developed to replicate robotic components of a lander manipulator system \cite{nayar2021development}. Cryogenic Ice Testing, Acquisition Development, and Excavation Laboratory (CITADEL) is developed to create icy moon representative thermal environments in order to enable sample acquisition and handling operations \cite{adams2020citadel}. 
Recent demonstrations also include autonomous sampling site selection \cite{ono2020ariel} and sampling autonomy \cite{bowkett2023demonstration} through field tests at terrestrial sites analogous to Europa. On the sampling system dynamics simulation and testing, JPL's Sampling Autonomy for Europa Lander Simulator (SAELSIM) \cite{bowkett2023demonstration} and NASA's Ocean Worlds Autonomy Testbed for Exploration Research and Simulation (OceanWATERS) provide simulation environment for autonomous lander operations development targeted towards ocean world environments \cite{catanoso2021oceanwaters}.    

To achieve stratified sampling of thin layers from the surface or subsurface of the icy moons or the plumes, high-fidelity perception of the sampling workspace is necessary for autonomy. The baseline goals for the perception system include: 
\begin{itemize}
    \item Recovering geometry (or) digital elevation model (DEM) of the sampling workspace.
    \item Recovering texture for sampling site selection.
    \item Sampling arm localization and fault detection.
    \item Excavation progress tracking.
\end{itemize}

But for systemic development of sensors or algorithms for sampling activities on Enceladus or Europa, high resolution ground truth of the surface geometry and composition is practically unavailable. Knowledge from flyby missions and telescopic observations do not yield the resolution scale required for sampling operations. Moreover, such data is not expected to be available until an orbiter or in fact an actual orbilander mission takes place. This makes the perception system development for the lander system very challenging.
Therefore, a rational development of candidate perception technologies necessitates high-fidelity simulations of the icy moon terrains that adhere to the current scientific observations.
In this paper, we put forward the challenges in simulating diverse icy moon environments and propose a Graphical Utility for Icy moon Surface Simulations (GUISS) for an engineering representation of the icy moon terrains. GUISS is developed as an empirically appropriate simulation tool based on an existing physically-based rendering engine with capabilities to simulate visual sensor modalities for high-fidelity rendering (see Figure~\ref{Fig:landerVisualization} for an example). 

In accordance with our motivation of investigating suitable perception systems for this task, we evaluate different algorithms for a stereo-based visual perception system using imagery generated by GUISS. In particular, we are interested in benchmarking classical and deep learning-based methods in visually inhospitable environments that contain highly reflective and low textured materials expected on Europa and Enceladus. 

%



\subsection{Paper Overview}

The paper is structured into seven sections. Following the introduction in Section 1, Section 2 provides a comprehensive look at the design requirements for a computer graphics tool aimed at synthetically visualizing the sampling workspace on icy moons. Section 3 offers an overview of pertinent research in the fields of planetary simulation and modeling.
In Section 4, the infrastructure and capabilities of the proposed graphical framework, GUISS, are described. Section 5 focuses on the performance evaluation of stereo depth estimation algorithms. Sections 6 and 7 delineate the conclusions and ongoing work, respectively.

%% file: Sections/02_DesignGoals.tex
\section{Study of Design Requirements} \label{sec:terrain_design_reqs}

Icy moons such as Europa and Enceladus, present unique challenges in simulating topographical and corresponding photometric properties. They house a sweeping diversity of geometric and optical features that needs to be simulated for lander operations  \cite{hand2017europa}. This section highlights our survey of geological characteristics of the said icy moons from the studies carried over the last two decades via flyby and telescopic observations. In the context of lander perception for sampling, this work is an effort towards development of a computer graphics software that can render versatile datasets of icy moon environments at a resolution apposite for sampling.

\subsection{Europa: Terrain Specifications}

The Europa Lander mission concept team at JPL has consolidated the current understanding of Europa's surface characteristics in the Terrain Specification Document \cite{cameron2022science}. Europa, a tectonically active icy celestial body roughly the size of Earth's moon, exhibits a high albedo ($\geq0.6$). With a mean surface temperature of $106$K at the equator, Europa's icy surface undergoes alteration due to high-energy radiation from Jupiter's magnetosphere through a process known as sputtering \cite{vorburger2018europa}.
Tidal flexing significantly influences Europa's surface features, contributing to the potential existence of a subsurface ocean beneath the icy shell. The stresses induced by tidal forces manifest as linear ridges, troughs, bands, fractures, and other chaotic features, reaching widths of up to $2$ km and heights of hundreds of meters \cite{pappalardo2013science}. These features, in conjunction with crater impacts, span thousands of kilometers across Europa's surface, alongside smooth regional plains scattered throughout the terrain.
Europa's surface primarily consists of water ice, predominantly in an amorphous form that gradually transitions into crystalline ice. Chemically, the presence of species such as salts and sulfates has also been identified \cite{hansen2004amorphous}.

\begin{figure}
\centering
\includegraphics[width=2.5in]{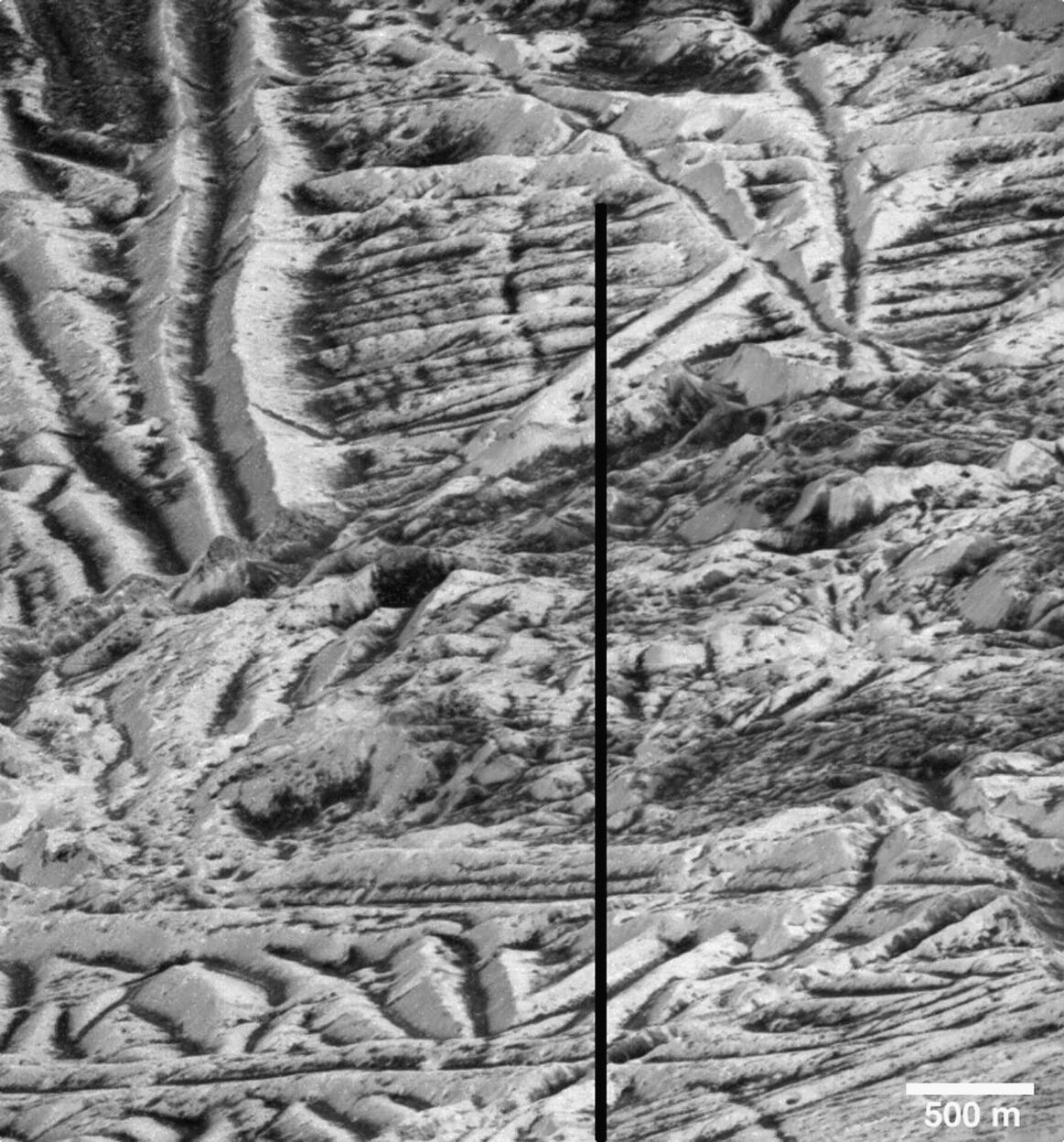}\\
\caption{This $6$ m/pixel is one the highest resolution images of Europa acquired by Galileo: Chaotic terrain with ridges in the foreground and background \protect\cite{pappalardo2013science}.}
\label{Fig:europa_HigRes}
\end{figure}

Surface composition is inferred largely from Galileo's Near Infrared Mapping Spectrometer (NIMS), Solid-State Imaging (SSI) camera, and Hubble Space Telescopic observations. 
A huge challenge for in-situ sampling investigations would be the lack of data on small-scale features of the Europan landscape. The highest resolution images of Europa from Galileo SSI are limited to $~6$ m/pixel in selected regions (see Figure  \ref{Fig:europa_HigRes}). Locally, the systematic albedo variations are deemed to be connected to the landform type, age as well as material disparities \cite{helfenstein1998galileo}. Young ridges with high topographic relief exhibit relatively high albedo. The ridges are separated by dark trough regions containing low-albedo ($0.17$) material.
It is reported that the low-albedo materials exhibit unusually intense shadow-hiding opposition effect. Conversely, areas with highly compacted particulates and coarse-grained regolith at topographically high landforms exhibit anomalously weak opposition surges \cite{helfenstein1998galileo}.

\subsubsection{Implications To Sampling Activities}

Based on current knowledge, Europa's topography consists of fractured and jagged regions that span from hundreds of meters to kilometers. For lander scale implications, this heterogeneity in the terrain is extrapolated to the sub-meter scale. The high-science interest targets for sampling have a substantial chance of extreme surface roughness at the centimeter scale \cite{pappalardo2013science}. 
Crystalline ice in the forms of snow and crushed ice are appropriate target materials to simulate for sampling activities \cite{catanoso2020analysis}. 
Strong opposition effects lead to extreme brightening of rough sampling surfaces under certain illumination phase angles.  
Diversity in surface textures and colors are attributed to compositional variations of water ice and hydrated materials, and result in differentiable shading effects.  
Although there are limitations on thermal or mechanical properties, sea-ice, glaciers, and permafrost are considered Earth analogues for Europan surface morphology \cite{preston2014planetary}.

\subsection{Enceladus: Terrain Specifications}

\begin{figure}
\centering
\includegraphics[width=2.5in]{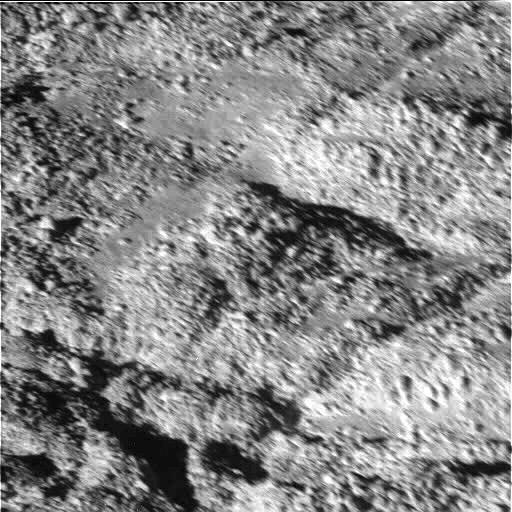}\\
\caption{This $4$ m/pixel is one the highest resolution images of Enceladus captured by Cassini's narrow angle camera: South polar terrain dominated by ice blocks \protect\cite{porco2006cassini}.}
\label{Fig:enceladus_HigRes}
\end{figure}

Enceladus is a small (approximately $500$ km diameter) but intriguing icy moon in the Saturnian system with direct evidence of an active hydrothermal system, subsurface ocean, and complex organic chemistry. With a geometric albedo of greater than $1.0$, it is the brightest of all objects in the solar system \cite{newman2008photometric}.
Flybys from Voyagers $1$ and $2$ and Cassini have shown that the surface of Enceladus is covered in pure water ice up to depths of $20-40$ kms on average \cite{postberg2009sodium}.
Cryovolcanic plume eruptions, from many individual jets along fractures known as Tiger Stripes in the South Polar Terrain (SPT), eject the subsurface ocean material onto Enceladus' surface and Saturn's E-ring \cite{kempf2008ring,helfenstein2015enceladus}. 

Surface morphology of Enceladus is a mix of smooth, craterless plains with numerous low-relief ridges in the leading hemisphere and heavily fractured corrugated plains in the younger SPT. 
Highest resolution image of Enceladus' Tiger Stripes region shows a large concentration of scattered ice boulders ($~10-100$ m across), surface disruptions ($2-10$ m high), faulting, and crumpling near the SPT with plume activity \cite{porco2006cassini,crow2015structural} (see Figure \ref{Fig:enceladus_HigRes}). 
Since its discovery by the NASA-ESA Cassini mission, Enceladus' plume is of high-scientific focus. 
Water ice grain composition of the plume particles has been confirmed through in-situ measurements. 
Investigation of Cassini's Visual and Infrared Spectrometer (VIMS) measurements revealed a scarcity of particles with sizes exceeding $3\,\mu$m at higher altitudes \cite{hedman2009spectral,dhingra2017spatially}. 
The particles falling from the plumes potentially contribute to the formation of an unconsolidated regolith layer and/or undergo sintering processes to form a solid ice layer \cite{kempf2010enceladus}.

The unusual photometric properties of Enceladus are attributed to its highly reflective surface layer covered with potentially fresh ice particle deposits ($<3 \,\mu$m) \cite{newman2008photometric}.
It is implied that the high geometric albedo is indicative of a highly backscattering surface with the absence of opaque material, unlike any natural frost layers observed on Earth \cite{buratti1988enceladus}.

\subsubsection{Implications To Sampling Activities}

Enceladus contains many sampling sites of significant scientific potential because of known high plume activity stretching to the orbit. 
For lander-based sampling activities, keen interest is on sites with exposed stratigraphy — in regions closest to the plumes where micron-sized fresh snow likely forms unconsolidated regolith layers \cite{kempf2008ring,kempf2010enceladus}. 
The collection space of interest also includes samples near the landing site, samples beneath the surface, and the top and depths of a reachable plume field. 
The terrain heterogeneities are extrapolated to sub-meter scales while maintaining plausible smooth reliefs mapped with a distribution of irregularly shaped ice blocks. 
Texturally, it was determined that ice particles with high amorphous quality are abundant in the SPT, while the reliefs in Tiger Stripes show a crystalline nature \cite{brown2006composition}.
With albedo differences of less than $20\%$, Enceladus' regolith exhibits multiple scattering and opposition effects in the visible spectrum \cite{buratti1988enceladus}. 
Visual perception and sampling autonomy are influenced by uncertainties in the surface reflection spectrum, illumination conditions, shape, size, and roughness of ice grains, as well as the presence of water vapor and subsurface CO$_2$ pockets in the lander vicinity, among others. 
Glacial plains with crevasses and subsurface oceans are considered Earth analogs for Enceladus' terrain morphology.

%% file: Sections/03_RelatedWork.tex
\section{Related Work}

\subsection{Planetary Surface Modeling}


High-fidelity image rendering and the perception sensor modeling for planetary surfaces has been pioneered in a number of developments. These are utilized as planning and visualization tools for relative navigation \cite{eapen2022narpa,bhaskara2023differentiable}, entry, descent and landing \cite{parkes2004planet}, and roving \cite{jain2020darts,bingham2023digital}, among others. 

PANGU (Planet and Asteroid Natural Scene Generation Utility) is being used at the European Space Agency for development and validation of vision based navigation systems targeting lander operations, surface relative navigation. and roving \cite{parkes2004planet}. PANGU enables creation of hierarchical (layered) landscapes and closed surfaces such as asteroids \cite{dubois2009testing}. It accommodates sensor models including camera, LiDAR, and RADAR for synthetic image generation. Integrated with SPICE kernels PANGU supports real-time closed-loop simulations. 
Similar to PANGU, Airbus' \textit{SurRender} software \cite{brochard2018scientific} and Texas A\&M's \textit{NaRPA} \cite{eapen2022narpa} are scientific rendering and sensor simulation tools optimized for space scene imaging.  
\textit{SurRender} additionally comes with modeling and data manipulation tools making it capable to handle large range of observing distances in a single instantiation \cite{brochard2018scientific}.
The Space Imaging Simulator for Proximity Operations (SISPO), based on Blender Cycles, is developed for space operations at Solar System scale and includes optical tools to simulate image aberrations as well as dust environments \cite{pajusalu2022sispo}.

JPL's DARTS (Dynamics Algorithms for Real-Time Simulation) framework is ubiquitously used in NASA missions for engineering quality simulations for planetary operations \cite{jain2003roams,jain2020darts,cameron2016dsends}. While focusing on simulation of multibody dynamics and real-time rendering for hardware-in-the-loop simulation of flight software, DARTS is also integrated with high-fidelity visual perception sensor models in its rendering pipeline \cite{aiazzi2022iris}. SAELSIM is a customization of DARTS, used for closed-loop simulation of Europa Lander's sampling mechanisms and tool-terrain interactions \cite{bowkett2023demonstration}. 

NASA's Digital Lunar Exploration Sites Unreal Simulation Tool (DUST) is catered towards graphical environment targeting surface mobility near the lunar south pole \cite{bingham2023digital}. DUST encapsulates the necessity for a specialized tool for ingesting and accurately modeling the planetary (lunar) terrain. Built on Unreal Engine 5, DUST leverages lunar datasets for modeling surface-level features, representative lighting conditions to generate lunar environment data products at scale.
Similarly, OAISYS \cite{muller2021photorealistic} offers a terrain simulation pipeline with the aim of providing training data of unstructured outdoor environments for robotic perception tasks such as instance segmentation and visual SLAM. The simulation is built on Blender and offers the capability to render imagery with varying geometry, texture, and illumination. BlenderProc \cite{Denninger2023} and VisionBlender \cite{cartucho2021visionblender} are other Blender based pipelines aimed at generating semantic datasets for the training of neural networks.     

Finally, OceanWATERS is a software simulation approach for modeling ocean world environment (landscapes and lighting) and lander robotic operations \cite{edwards2021autonomy}. With focus on simulation of lander dynamics, OceanWATERS uses OGRE rendering engine for visualization for realism and openGL for fast rendering. The platform is equipped with synthetic terrain generation for icy worlds with separate treatment for near and far field terrain features. However, systematic treatment of surface textures and reflectance models are claimed to be under development.

The rendering framework proposed in this paper represents an advancement aimed at enhancing the scientific understanding of icy world landscapes to engineer a realistic representation of sampling sites on icy moons. This pipeline is designed to model the surface topography, texture, and illumination of icy moons (specifically Europa and Enceladus) in accordance with scientific observations. The goal is to create a versatile dataset for vision-based sampling on icy moons, encompassing the geometric and optical diversity of potential sampling workspaces.

This paper explores the capabilities of procedural terrain and texture generation under representative illumination conditions. The rendering framework, GUISS, capitalizes on the capabilities of Blender Cycles for high-fidelity image synthesis. We augment the icy world data assets using procedural generation methods and reconstructions from Earth analogues. The performance of state-of-the-art depth estimation pipelines is evaluated for perceptually challenging visual imagery of icy moons.





%% file: Sections/04_Pipeline.tex
\section{Rendering Infrastructure}\label{sec:rendering_capabilities}

As indicated, a versatile visual simulation environment is indispensable in order to design and validate perception algorithms for icy moon surface operations.
This is especially applicable for sampling activities on the icy worlds where the lander mission has to be designed almost unknowing of the geometric and photometric properties of the sampling workspace.
The {Graphics Utility for Icy Moon Surface Simulations (GUISS)} has been developed to provide a modular pipeline for physically based rendering of representative icy world surfaces. 
This capability is harnessed for generating versatile datasets that cover the spectrum of perceptually challenging conditions on icy worlds. These synthetic datasets, spanning expected visual conditions, are essential for training and testing computer vision algorithms intended for sampling autonomy.

GUISS is a Python-based software that utilizes Blender Cycles for physically based path tracing. Blender's path-tracer accurately simulates the physics of light-surface interactions in a scene to produce photorealistic renders with global illumination effects. The Cycles engine supports various shader types, including (i) Principled bidirectional scattering distribution function (BSDF), a versatile mix of shading parameters into a single node; (ii) Transparent and Glass shaders for rendering materials with transparency and refraction; (iii) volume shader for simulating complex atmospheric effects such as snow; and (iv) emission shader for creating self-illuminating surfaces. Additionally, through Open Shading Language (OSL), Cycles enables the creation of custom shaders, such as the Hapke bidirectional reflectance distribution function (BRDF).

\subsection{Overview}

\begin{figure}
\centering
\includegraphics[width=3.35in]{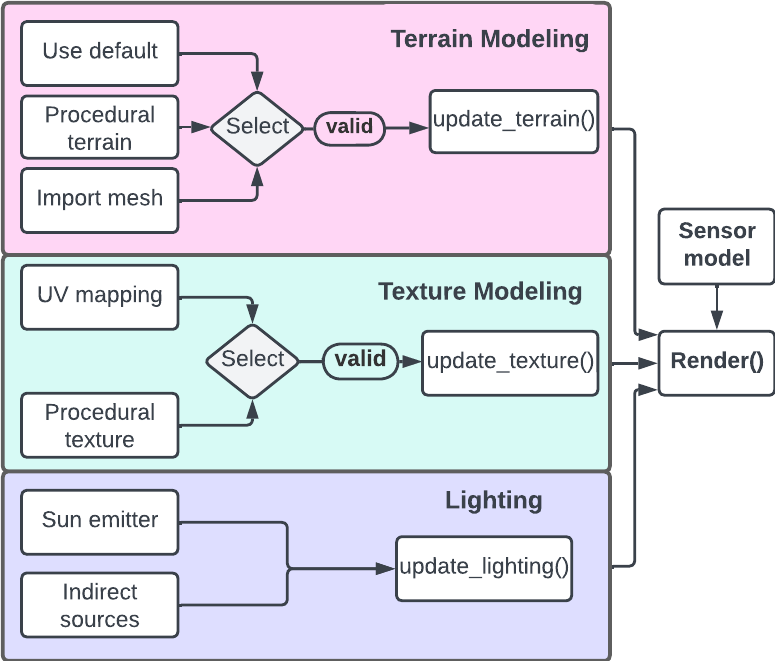}
\caption{{Overview of the pipeline of operations: synthetic data assets for the icy moon surfaces are created from modeling of terrain, texture, and lighting elements. For chosen sensor parameters, Blender Cycles is utilized to render stereo imagery.}}
\label{Fig:rendering_pipeline}
\end{figure}

The rendering infrastructure is designed to provide modularity, allowing users to configure virtual scene geometry, textures, lighting conditions, and sensor parameters through a Python interface. 
Existing features are reconfigurable, and additional models can be seamlessly integrated into the framework. 
To set up a virtual scene, the rendering pipeline consists of three main building blocks: (i) terrain modeling, (ii) texture modeling, and (iii) lighting setup. 
A high-level overview of the rendering pipeline is illustrated in Figure \ref{Fig:rendering_pipeline}. Synthetic images are generated by modeling the camera intrinsic parameters of either monocular, stereo, or depth imagers. 
The framework transfers the environment and sensor model description to Blender Cycles for producing output data products.

\subsection{Terrain Modeling}

Following the theoretical arguments described in Section \ref{sec:terrain_design_reqs}, GUISS facilitates the development and utilization of a broad range of observation geometries. Three different procedures are adopted to handle various physical geometries of configurable roughness. Blender's capabilities are integrated to create or modify terrain models to the desired scale, shape, and resolution.

In the first default method for the software, GUISS is packaged with a limited set of meshes that are manually sculpted and built to represent the topography of rugged and glacier-like terrains. These meshes are generated from procedural methods but include hand-crafted features for additional details. The default meshes offer a starting point for off-the-shelf rendering of representative scenes and serve as a foundation for further customization using Blender. Figure \ref{Fig:landerVisualization} shows an example of default terrain (with rock features) rendered using GUISS.

\ignore{
\begin{figure}
\centering
\includegraphics[width=3in]{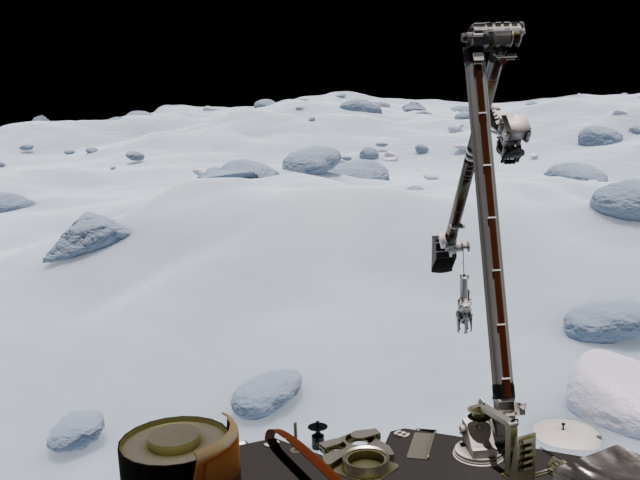} 
\caption{Lander arm visualization of representative icy-moon terrain rendering using {GUISS}.}
\label{Fig:landerVisualization}
\end{figure}
}


\subsubsection{Importing Landscapes}

\begin{figure*}[t]
\centering
\includegraphics[width=0.24\textwidth]{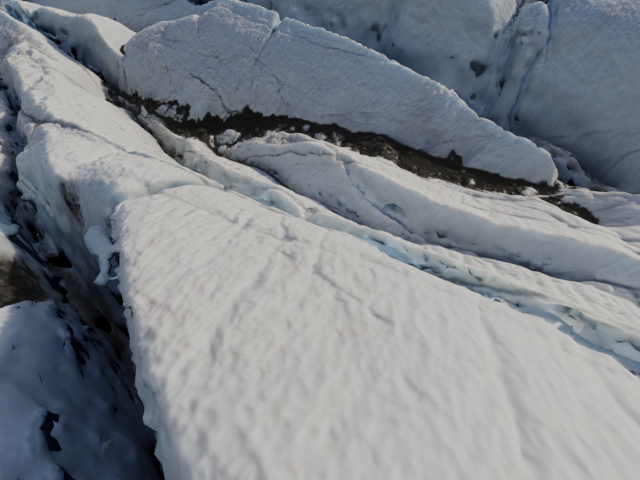}
\includegraphics[width=0.24\textwidth]{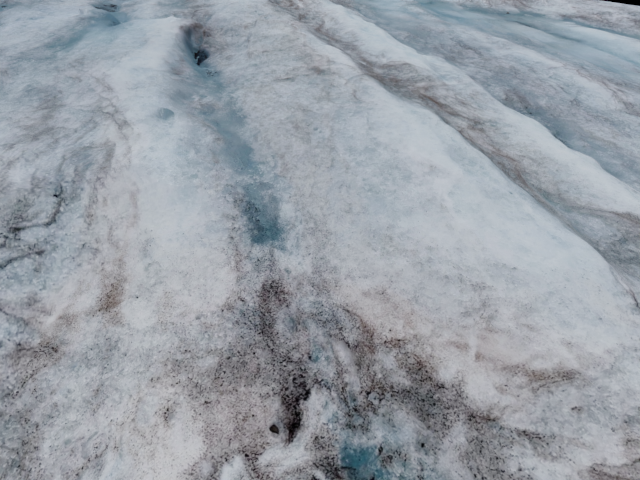}
\includegraphics[width=0.24\textwidth]{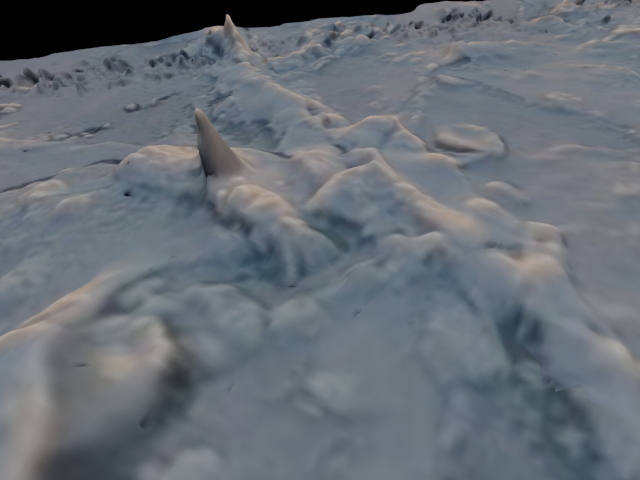}
\includegraphics[width=0.24\textwidth]{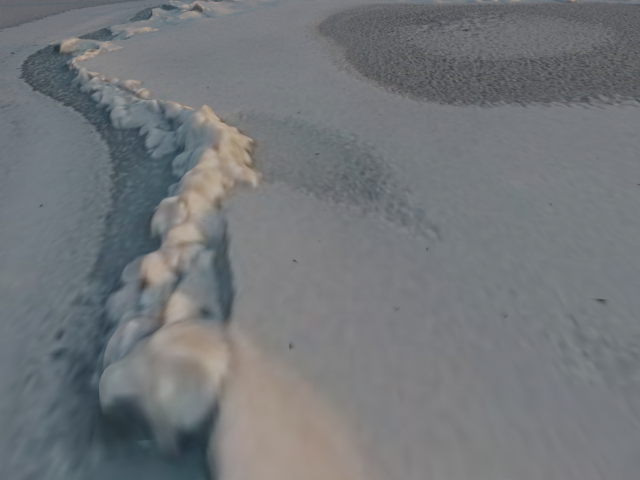}
\\
\includegraphics[width=0.24\textwidth]{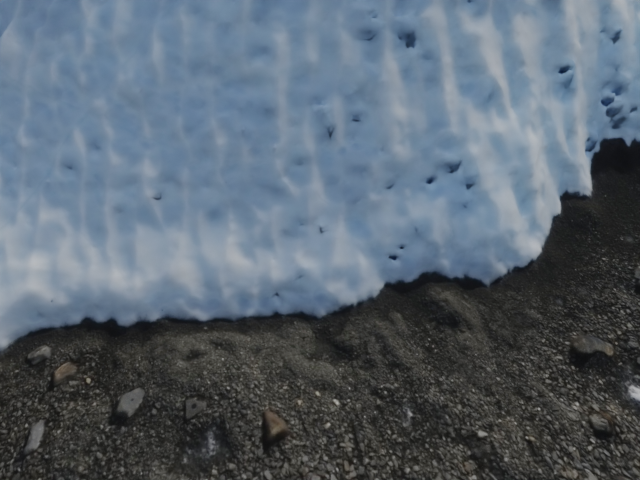}
\includegraphics[width=0.24\textwidth]{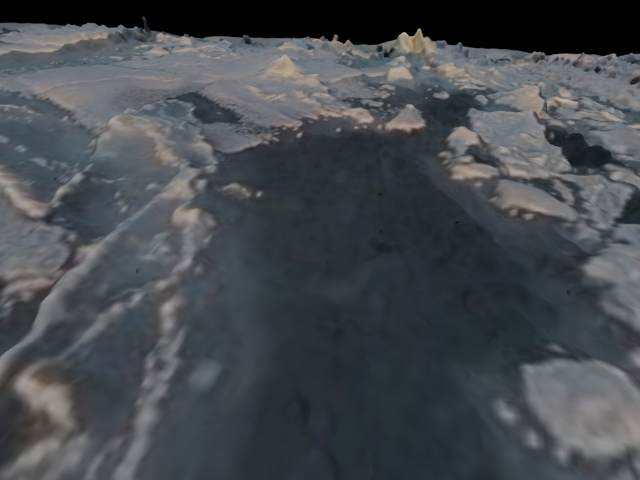}
\includegraphics[width=0.24\textwidth]{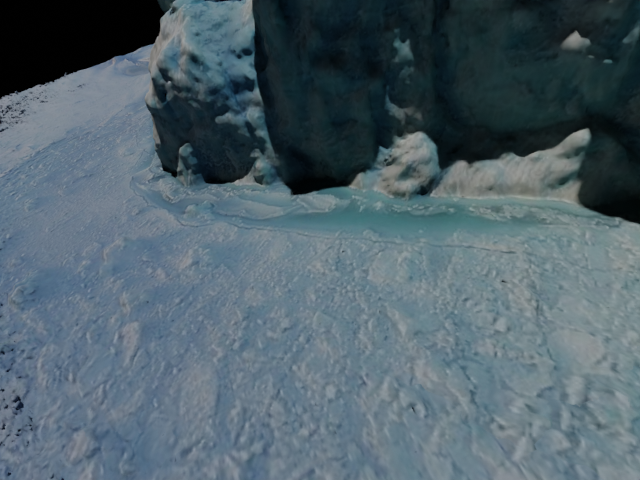}
\includegraphics[width=0.24\textwidth]{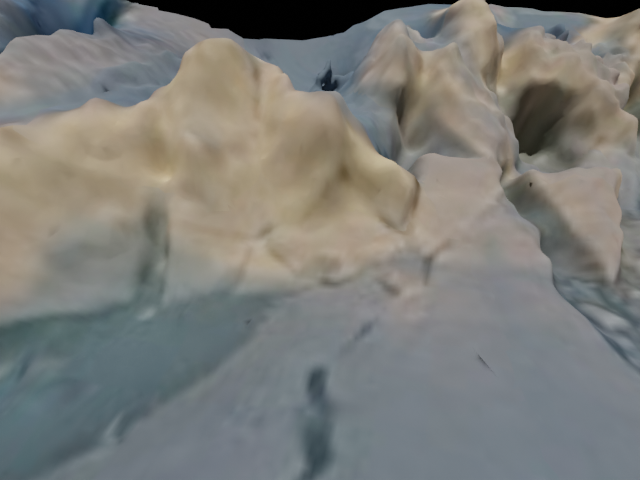}
\caption{Renderings of scenes reconstructed from Europa lander field testing campaign at Matanuska Glacier, Alaska.}
\label{fig:examples_scene_reconstructions}
\end{figure*}

GUISS supports the sourcing of terrain data from various origins and provides utilities to modify them. Topographic data can be imported in a range of formats supported by Blender for virtual scene generation. This feature facilitates working with meshes reconstructed from icy-moon exploration test campaigns on Earth analogues. Figure \ref{fig:examples_scene_reconstructions} displays virtual analogs from field testing campaigns for the Sampling Autonomy for Europa Lander (SAEL) project at Matanuska Glacier, AK, USA. The renders showcase real reconstructed models of glacial terrain. Similarly, meshes designed in specialized terrain design applications can be easily imported as landscapes (see Figure \ref{Fig:Gaea_procedural}).



\subsubsection{Procedural Terrain Generation}

Procedural terrain generation is a geometric technique in which terrain features are systematically generated from randomness using 3D basis functions \cite{perlin1985image,worley1996cellular}. Noise models serve as the foundation for the procedural synthesis of mesh nodes, which are then smoothly interpolated to create a continuous and naturally-looking surface. These methods often leverage fractal noise, a mathematical construct created by the summation of multiple scales (or octaves) of noise basis functions.

\begin{figure}
     \centering
     \begin{subfigure}[b]{0.23\textwidth}
         \centering
         \includegraphics[width=\textwidth]{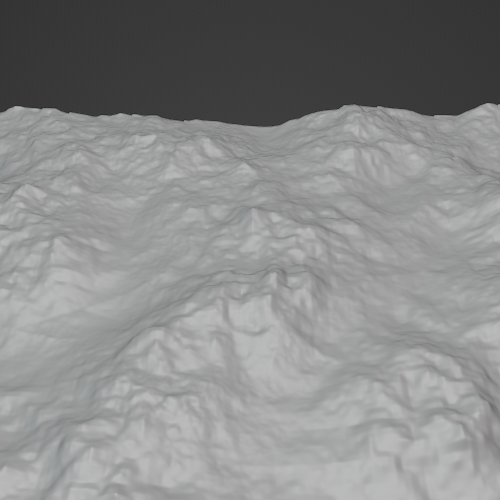}
         \caption{Terrain of low ruggedness.}
         \label{fig:procedural1_low}
     \end{subfigure}
     \hspace{2mm}
     \begin{subfigure}[b]{0.23\textwidth}
         \centering
         \includegraphics[width=\textwidth]{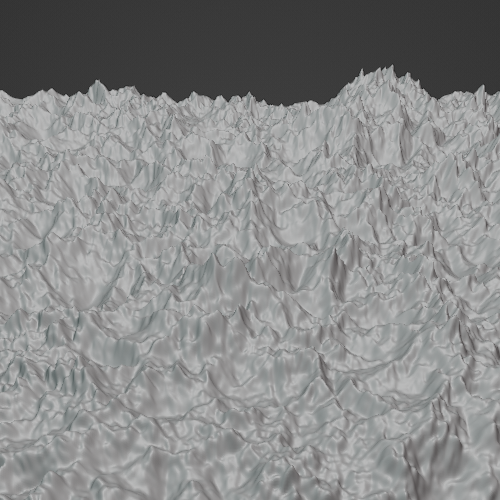}
         \caption{Moderately rugged terrain.}
         \label{fig:procedural1_med}
     \end{subfigure}
    \hfill
     \begin{subfigure}[b]{0.23\textwidth}
         \centering
         \includegraphics[width=\textwidth]{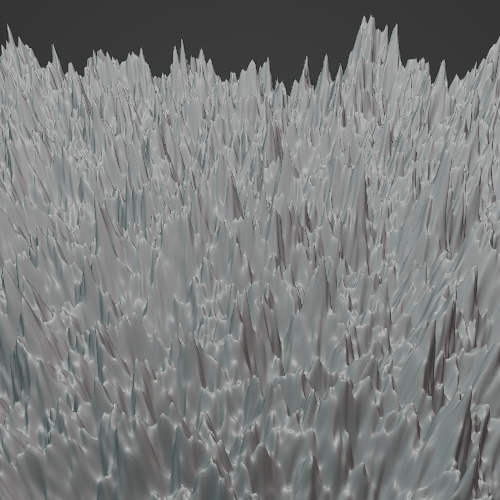}
         \caption{Highly rugged terrain.}
         \label{fig:procedural1_high}
     \end{subfigure}
     \hspace{2mm}
     \begin{subfigure}[b]{0.23\textwidth}
         \centering
         \includegraphics[width=\textwidth]{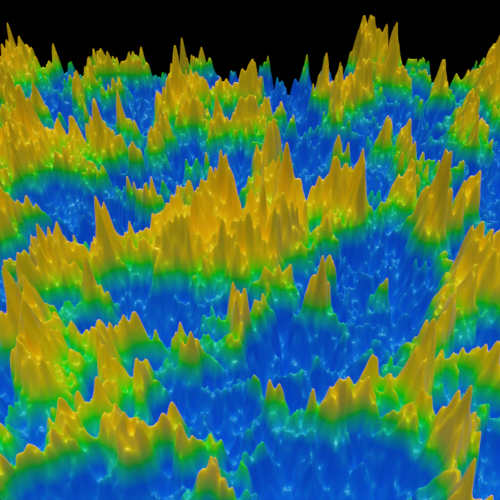}
         \caption{Icy penitente features.}
         \label{fig:procedural1_penitente}
     \end{subfigure}
        \caption{Top views of multifractal terrains synthesized from procedural methods using {GUISS}. In bottom right the icy penitente features reach up to $6$m high.}
        \label{Fig:terrain_procedural}    
\end{figure}

\begin{figure}
     \centering
     \begin{subfigure}[b]{0.23\textwidth}
         \centering
         \includegraphics[width=\textwidth]{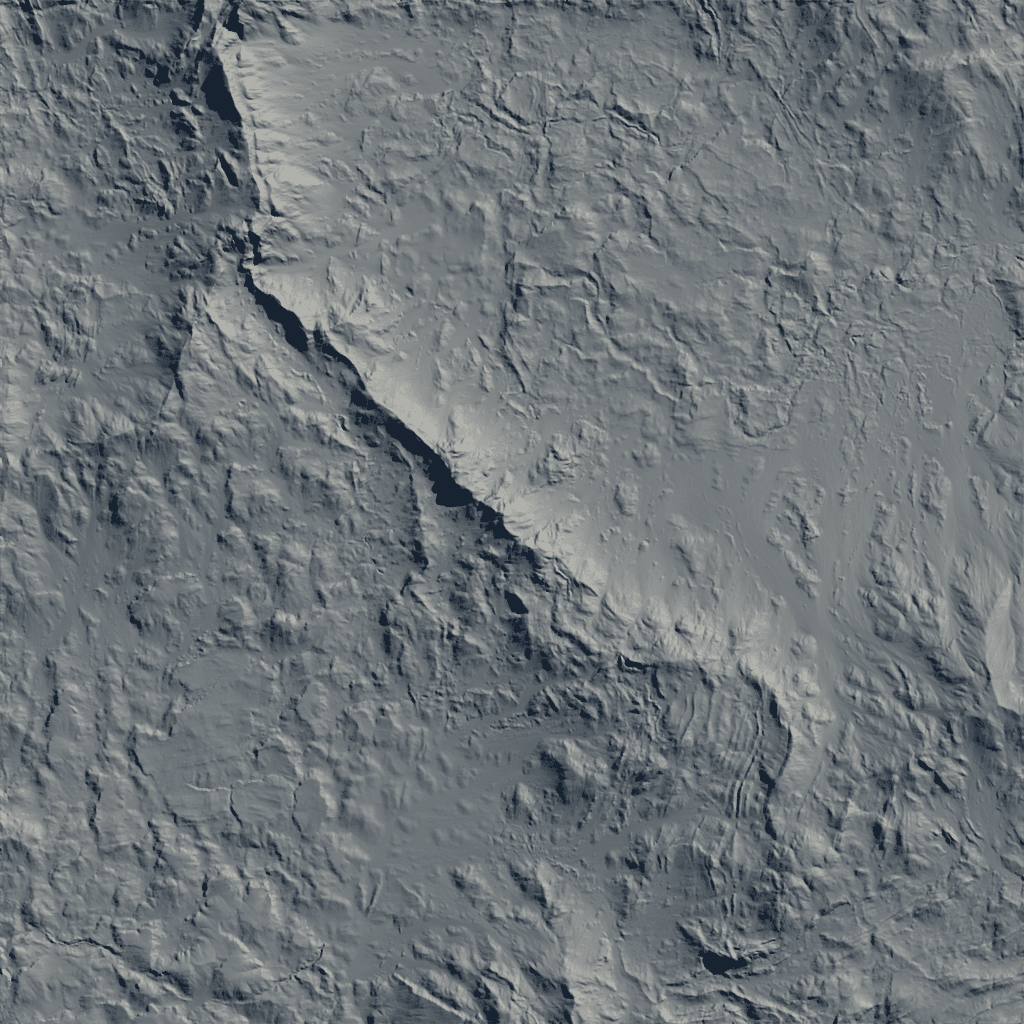}
         \caption{Rocky terrain with low ridges.}
         \label{fig:procedural1}
     \end{subfigure}
     \hspace{2mm}
     \begin{subfigure}[b]{0.23\textwidth}
         \centering
         \includegraphics[width=\textwidth]{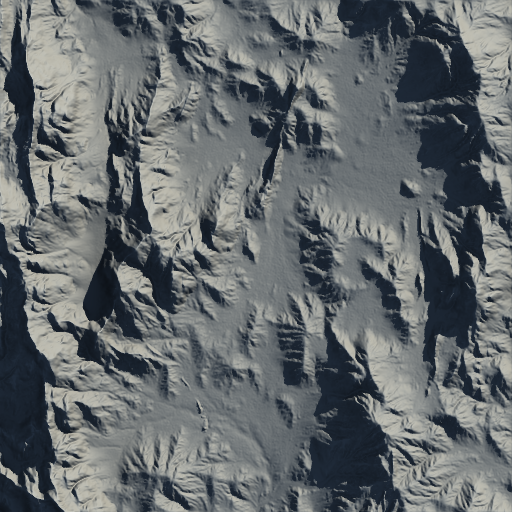}
         \caption{Terrain with high concentration of ridges.}
         \label{fig:procedural2}
     \end{subfigure}
        \caption[Procedural terrains generated on Quadspinner Gaea application]{Procedural terrains generated on Quadspinner Gaea application {\cite{QuadspinnerGaea}}.}
        \label{Fig:Gaea_procedural}
\end{figure}

Noise basis functions such as Perlin noise, Voronoi noise, midpoint displacement, turbulence, and others are combined to generate multi-fractal patterns and complex terrains \cite{ebert2003texturing}. Adjusting the amplitudes and scales for individual noise functions allows for the generation of terrains with varying levels of detail and complexity. Therefore, procedural generation is a valuable tool for synthesizing natural surfaces without requiring hand modeling of geometric details. It is not only applicable for generating an entire mesh but also for increasing the details in existing meshes, offering a robust and flexible way to create versatile 3D assets. 
Examples are highlighted in Figures \ref{Fig:terrain_procedural} and \ref{Fig:Gaea_procedural}. In Figure \ref{Fig:terrain_procedural}, geometry is procedurally synthesized by varying noise amplitude to attain ruggedness of varying scales. Spike-like icy structures called penitentes can be synthesized using procedural methods. Figure \ref{Fig:Gaea_procedural} shows multifractal terrains interactively created using various noise basis functions on the Quadspinner Gaea terrain design application.


\subsubsection{Rock distribution}

\newcommand*\circled[1]{\tikz[baseline=(char.base)]{
            \node[shape=circle,draw,inner sep=2pt] (char) {#1};}}
            
\begin{figure}[t]
\centering
\begin{tikzpicture}
    \node[anchor=south west, inner sep=0] (image) at (0,0) {\includegraphics[width=0.46\textwidth]{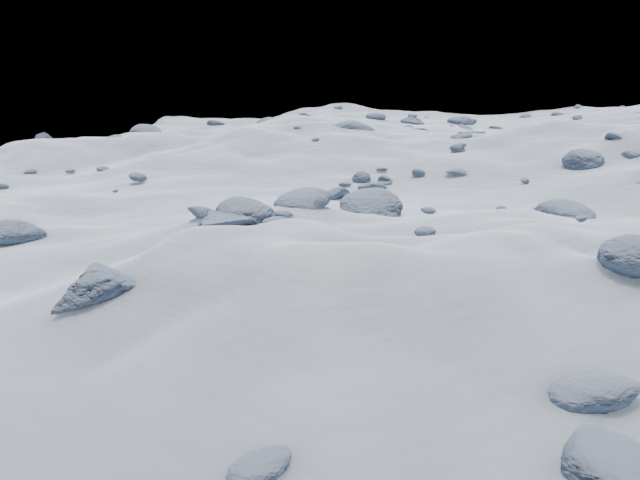}};
    \coordinate (marker) at (0.5,5.5); 
    \draw[white, fill=white, draw] (marker) node { \textbf{ \circled{1} }   };
  \end{tikzpicture}
\hfill
\\
\begin{tikzpicture}
    \node[anchor=south west, inner sep=0] (image) at (0,0) {\includegraphics[width=0.23\textwidth]{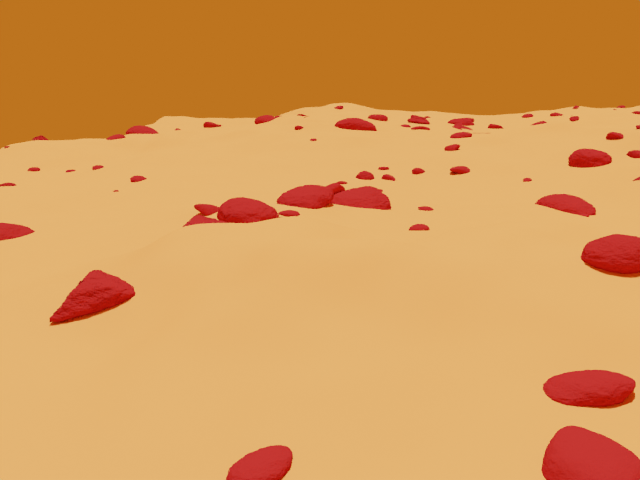}};
    \coordinate (marker) at (0.5,2.6); 
    \draw[white, fill=white, draw] (marker) node { \textbf{ \circled{2} }   };
\end{tikzpicture}
\begin{tikzpicture}
    \node[anchor=south west, inner sep=0] (image) at (0,0) {\includegraphics[width=0.23\textwidth]{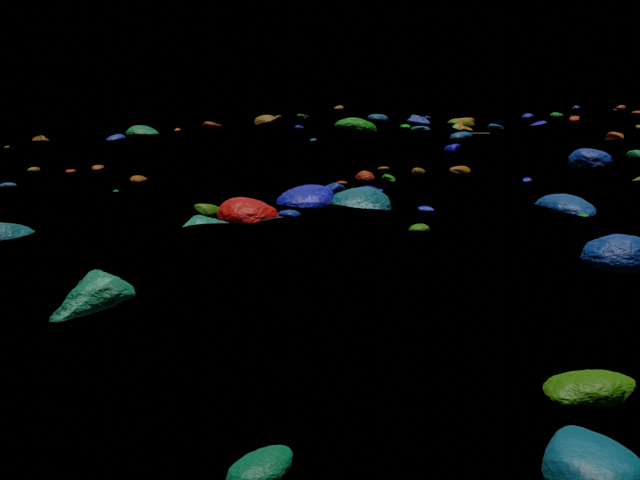}};
    \coordinate (marker) at (0.5,2.6); 
    \draw[white, fill=white, draw] (marker) node { \textbf{ \circled{3} }   };
\end{tikzpicture} 
\caption{Example of rock distribution with varying scales and densities. {{\textcircled{\small 1}}} is the RGB render of the representative icy-world environment, {\textcircled{\small 2}} and {\textcircled{\small 3}} are the corresponding semantic and instance segmentation masks, respectively.}
\label{fig:examples_rocks}
\end{figure}


\ignore{
\begin{figure}
\centering
\includegraphics[width=3.25in]{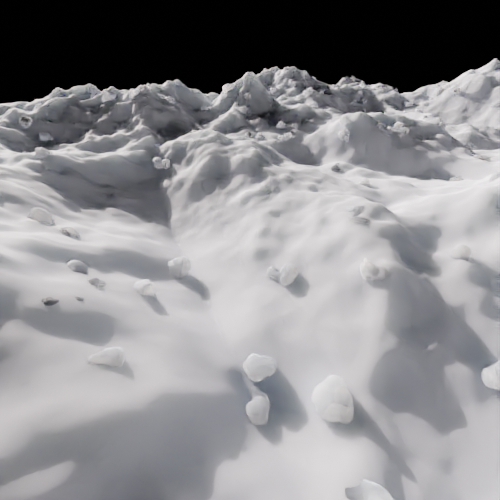}\\
\caption{{Rock distribution on a synthetic, representative icy-world environment.}}
\label{Fig:rockDistribution}
\end{figure}
}

To account for the low resolution features of icy moons (Enceladus terrain in Figure \ref{Fig:enceladus_HigRes}) observed from orbit and to increase the realism, the capability to add a random distribution of rocks has been incorporated. 
The distribution density, scale, shape, and orientation of rocks are user-configurable for generating icy rocks at runtime. The rock distribution is primarily for visual features, interpolating from the presence of $10-100$ m rocks on Enceladus' South Polar Terrain image ($20\times20$ km) to $10-100$ cm (across) rocks distributed across the sampling workspace ($20\times20$ m).
GUISS introduces texture to the rocks, where their albedo corresponds to the albedo of the terrain but can also be configured to exhibit different surface properties. Example is shown in Figure \ref{fig:examples_rocks}. 
Moreover, the figure also shows Blender's synthetic data generation capabilities extended to provide annotations for training of neural networks.
Semantic segmentation masks are generated using indices of object instances in the scene. 
Instance masks are created by assigning diffuse material colors to each object instance followed by a rendering pass. 
Pending scientific observations at this point, the rock distribution is not statistically representative of icy worlds. Nevertheless, the software is capable of adapting to different configurations for versatility and realism.


\subsection{Texture Modeling}

Scientific inquiries into the surface texture properties of icy moons cover a broad spectrum of considerations, as detailed in Section \ref{sec:terrain_design_reqs}. These properties include the intricate details of the surface regolith internal structure, compositional variations, albedo, and roughness. To effectively simulate engineering representations of these icy bodies, we rely on scientific hypotheses that delineate the types of deposits present on moons like Europa and Enceladus. The rendering surface texture capabilities are continuously evolving in the presented framework. As new data becomes available, the software adapts to incorporate new learnings, improving fidelity and visual acuity.

To this extent, we employ two methods for modeling textures on the icy-moon terrain: (i) texture mapping and (ii) procedural displacement textures, or a combination of both.


\subsubsection{Texture Mapping}

Texture mapping onto geometric terrain poses a critical challenge, playing a pivotal role in engineering simulations for icy moon terrains. The primary objective is to faithfully represent the high albedo of icy moon surfaces and capture intricate texture details that reflect surface roughness realistically. In our approach, Blender's shader nodes are utilized for UV texture mapping. However, due to disparities in terrain resolutions and texture scales, traditional and direct UV mapping can result in a repetitive, tiled appearance.
To address this challenge, we implement a workaround by mapping the object coordinates of the terrain to the texture UV coordinates. This technique, also used in the DUST software for lunar surface textures \cite{bingham2023digital}, helps mitigate the issue of inconsistent UV scaling. Additionally, to minimize apparent repetition and ensure a smoother texture response, we employ a blending approach that seamlessly merges repetitive texture tiles for enhanced visual continuity.

\begin{figure}[t]
\centering
\includegraphics[width=0.23\textwidth]{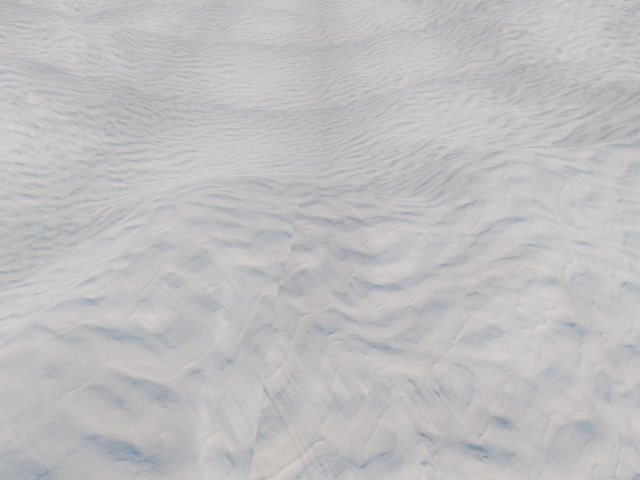}
\includegraphics[width=0.23\textwidth]{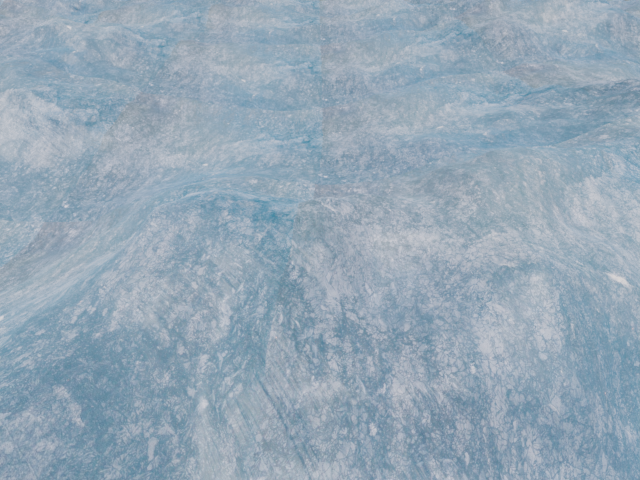}
\\
\includegraphics[width=0.23\textwidth]{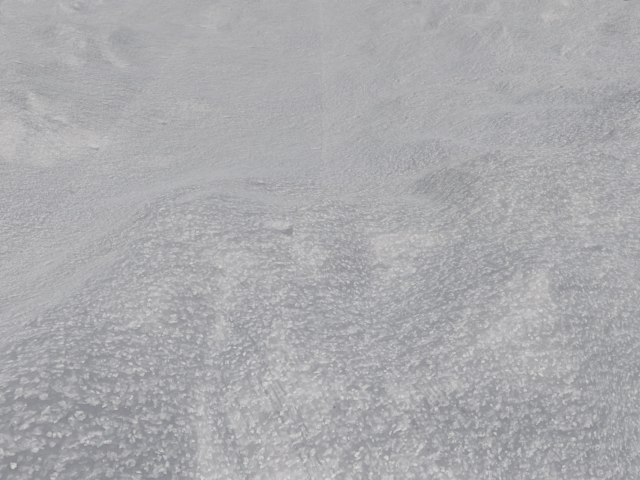}
\includegraphics[width=0.23\textwidth]{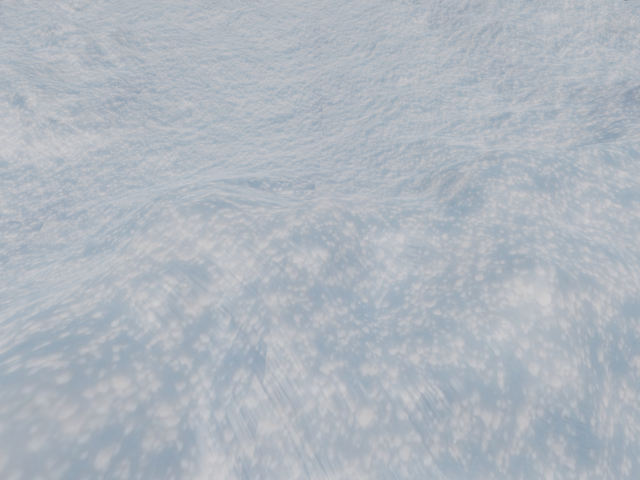}
\\
\includegraphics[width=0.23\textwidth]{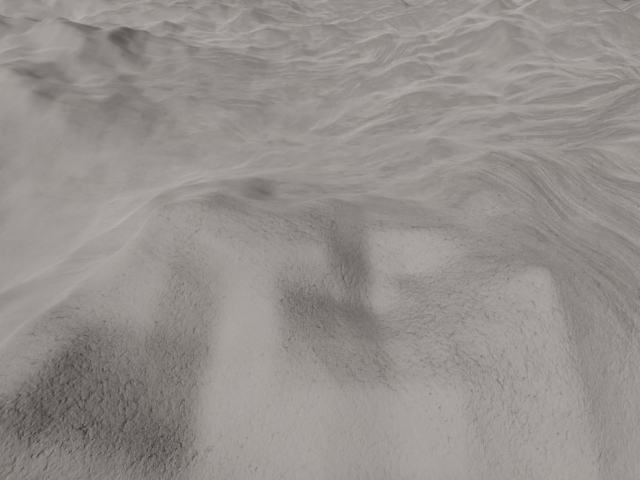}
\includegraphics[width=0.23\textwidth]{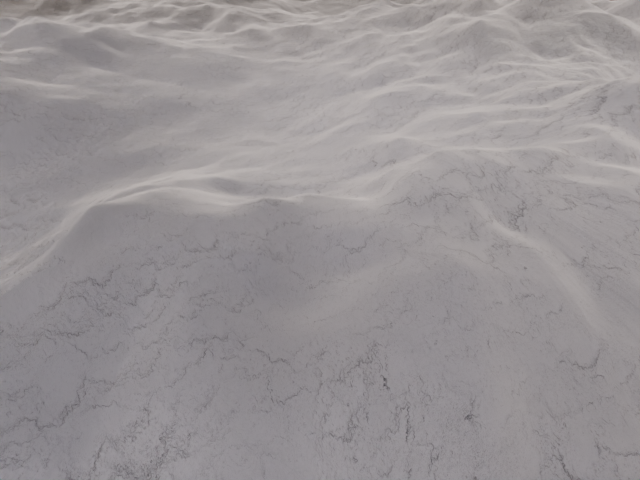}
 \caption{UV texture mapping: Example renders showing UV mapping of texture images onto a representative mesh.}
\label{fig:examples_texture_variation}
\end{figure}

Surface albedo maps define the base color of the terrain. In our texture mapping approach, albedo maps depicting high reflectance, characteristic of snowy or ice-like surfaces, are applied for UV mapping. Using Blender's Principled BSDF shader node, transmission, displacement, normal, and roughness may also be mapped similar to the albedo or image texture maps. The goal is to compile a dataset featuring finely detailed texture maps that reflect icy moon surface albedo at both orbital and lander distances. Figure \ref{fig:examples_texture_variation} shows an example of terrain rendered using UV mapping. The image textures and albedo maps are collected from publicly available sources.

\begin{figure*}[t]
\centering
\includegraphics[width=0.3\textwidth]{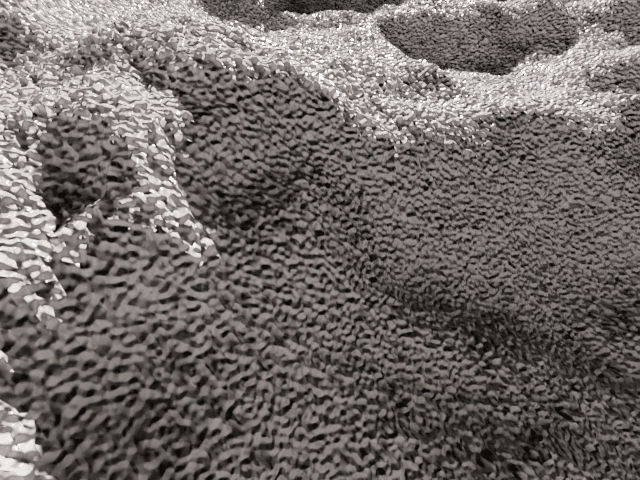}
\includegraphics[width=0.3\textwidth]{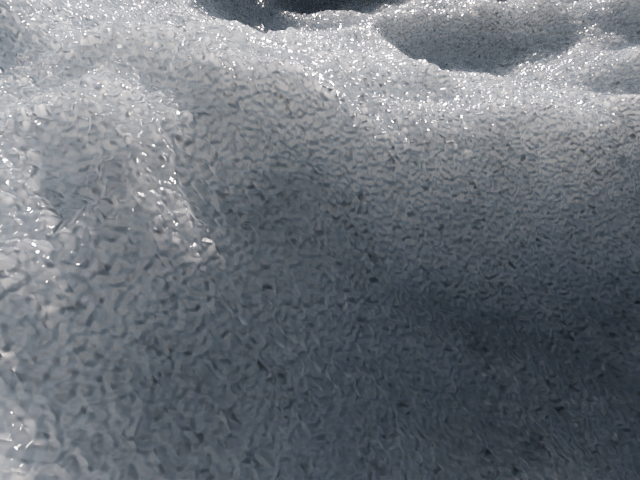}
\includegraphics[width=0.3\textwidth]{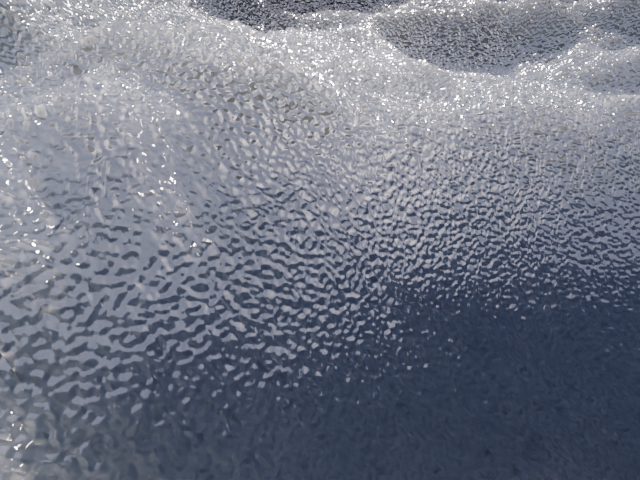}
\\
\includegraphics[width=0.3\textwidth]{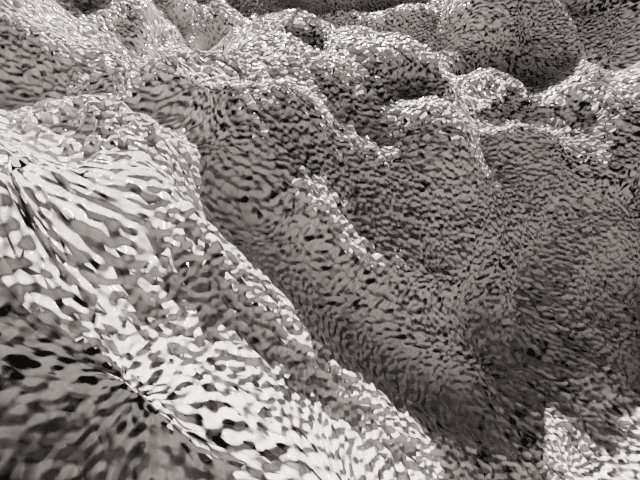}
\includegraphics[width=0.3\textwidth]{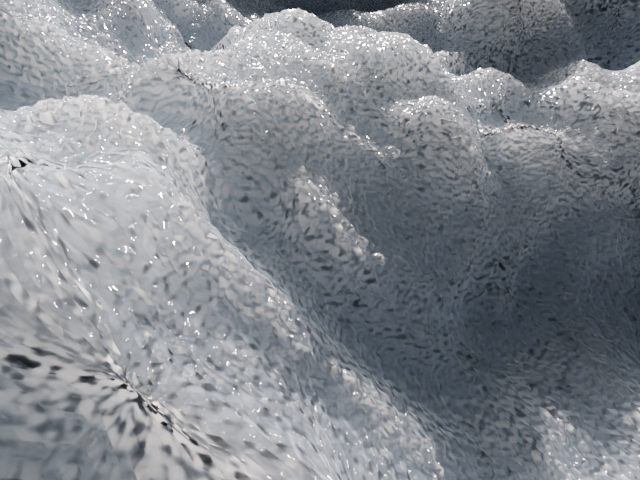}
\includegraphics[width=0.3\textwidth]{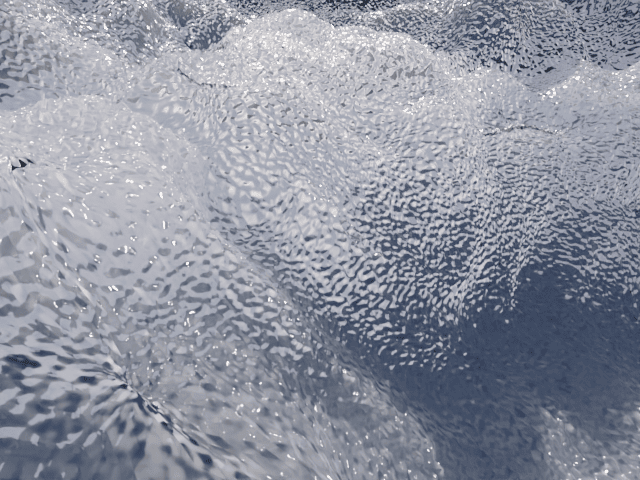}
\\
\includegraphics[width=0.3\textwidth]{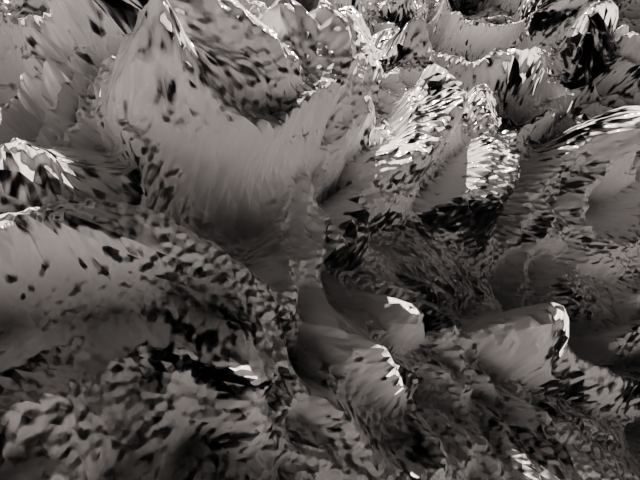}
\includegraphics[width=0.3\textwidth]{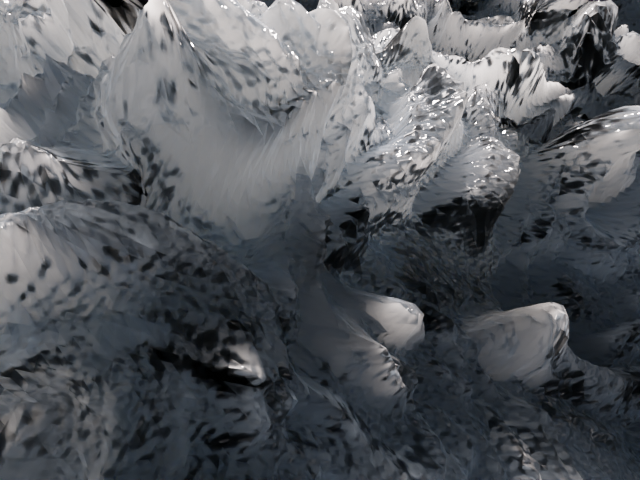}
\includegraphics[width=0.3\textwidth]{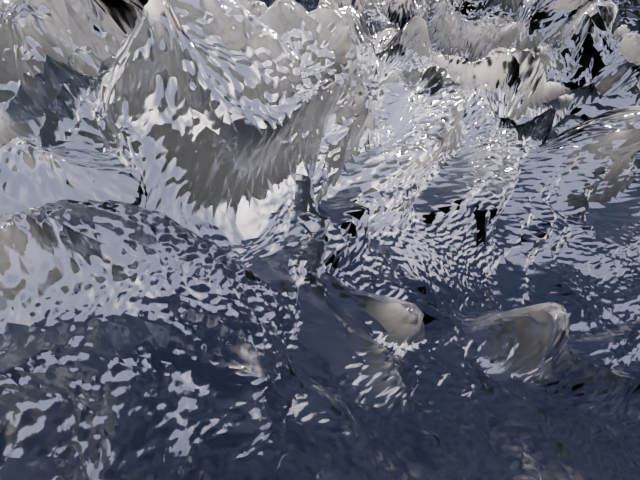}
\\
\includegraphics[width=0.3\textwidth]{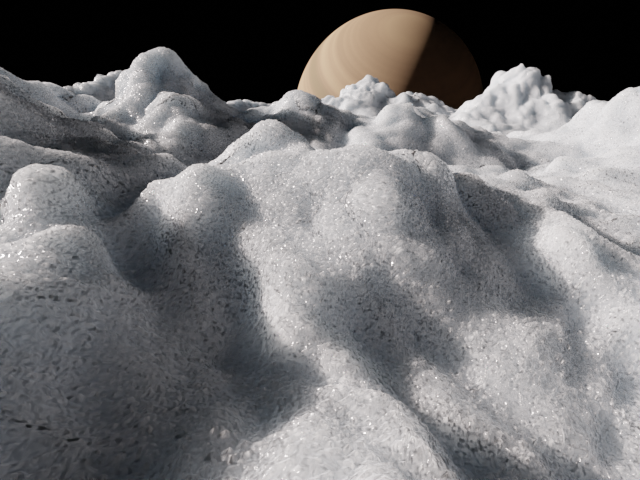}
\includegraphics[width=0.3\textwidth]{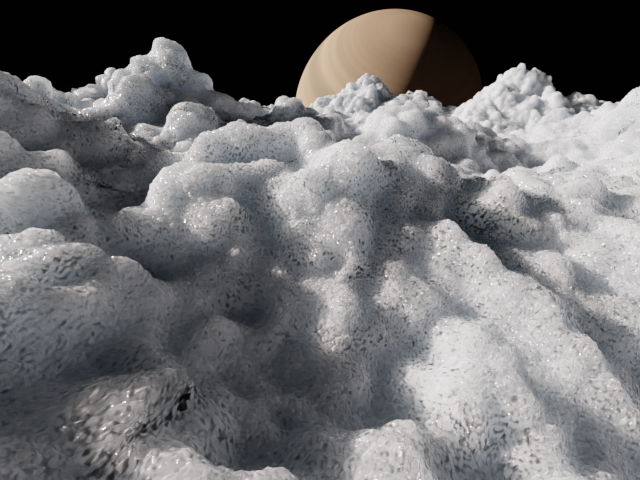}
\includegraphics[width=0.3\textwidth]{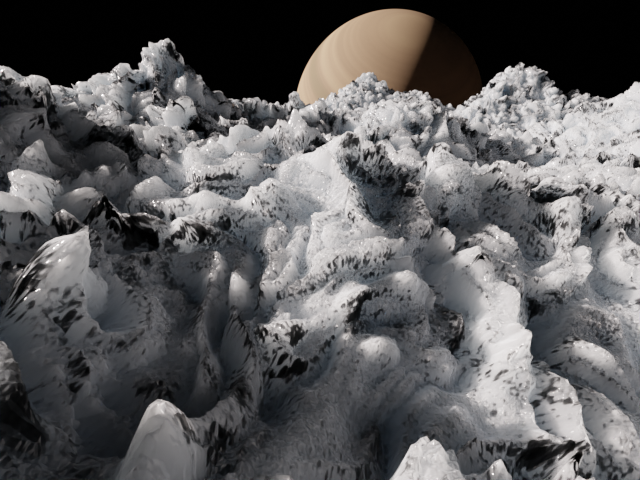}
\caption{First three rows from left-to-right: Increasing subsurface factor and transmission of the snow material. Bottom row from left-to-right: Increasing procedural texture noise.}
\label{fig:examples_generative_subsurface}
\end{figure*}


\ignore{
\begin{figure}
\centering
\includegraphics[width=3.25in]{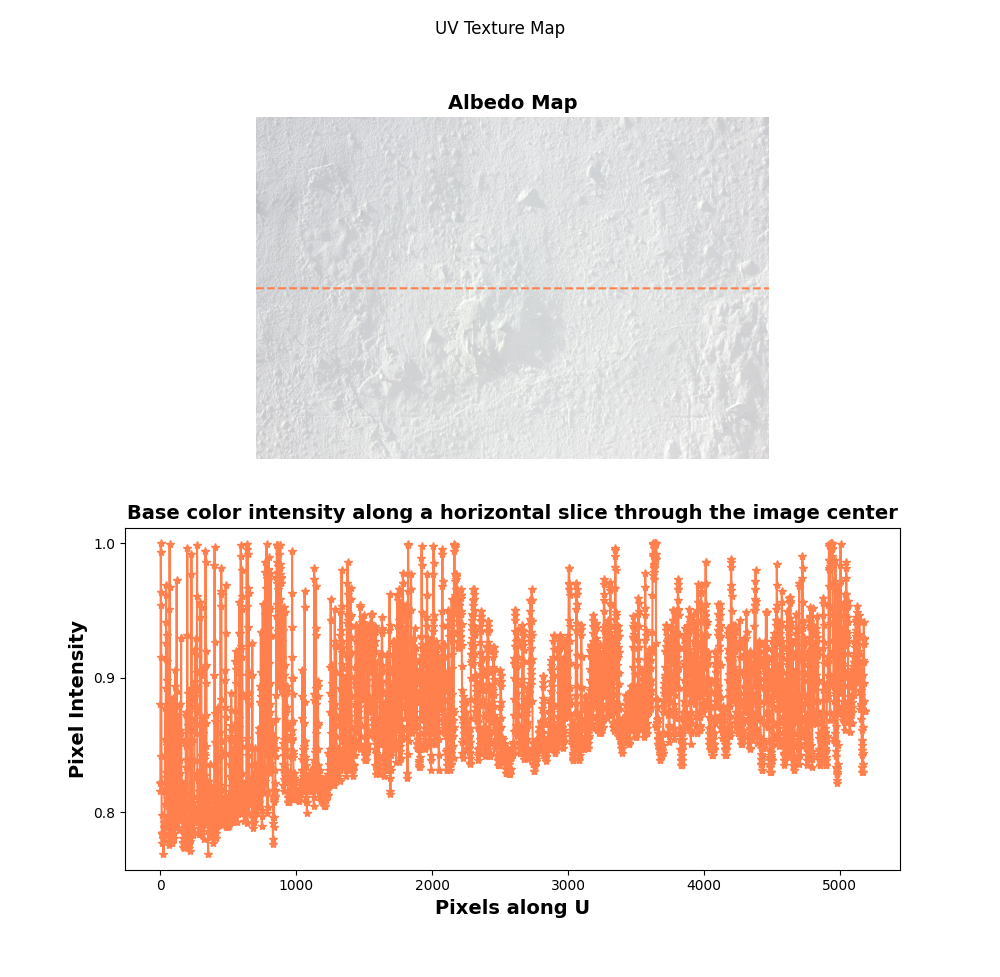}\\
\caption{Texture mapping for a synthetic Enceladus environment.}
\label{Fig:albedoMap}
\end{figure}
}

\subsubsection{Procedural Texture Generation}

Procedural texturing shares a conceptual similarity with procedural terrain generation, employing a textured approach to modify the geometry of an existing mesh. In contrast to procedural terrain generation, which involves creating entirely new landscapes, procedural texturing focuses on refining and enhancing the details of an existing mesh. This process involves the displacement of the upper regolithic surface of the mesh, introducing intricate variations through changes in surface normals.

Within the Blender Cycles framework, not only can displacement effects be achieved, but the platform also facilitates the creation of procedural adjustments to surface roughness and normals. Leveraging a combination of iterated noisy textures, Blender Cycles enables the generation of textured surfaces that authentically represent phenomena such as realistic snow textures with crystalline material-like reflections and subsurface scattering, as visually demonstrated in Figure \ref{fig:examples_generative_subsurface}.



\begin{figure}[t]
\centering
\includegraphics[width=0.15\textwidth]{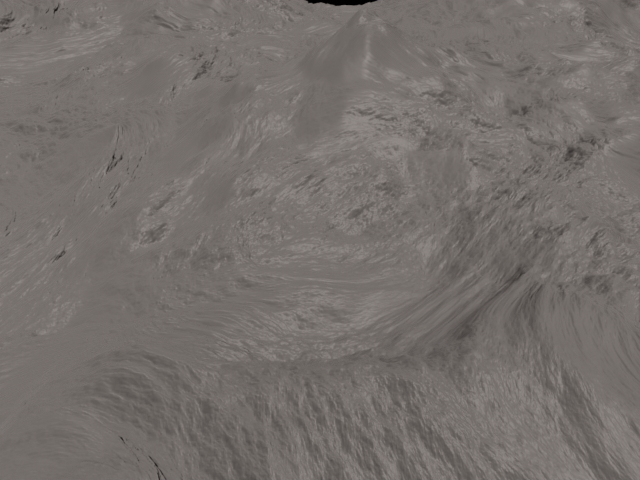}
\includegraphics[width=0.15\textwidth]{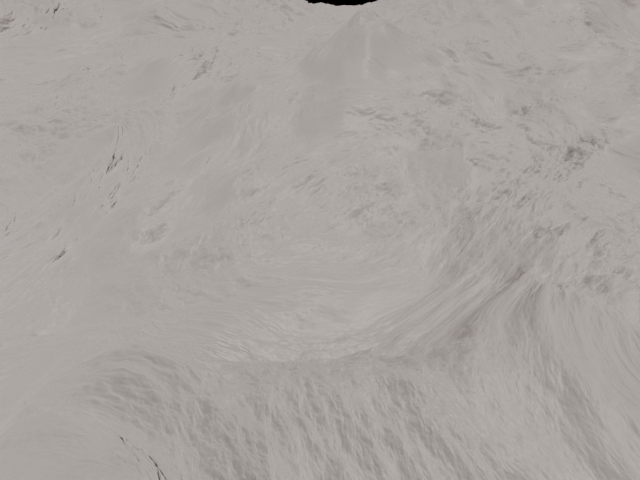}
\includegraphics[width=0.15\textwidth]{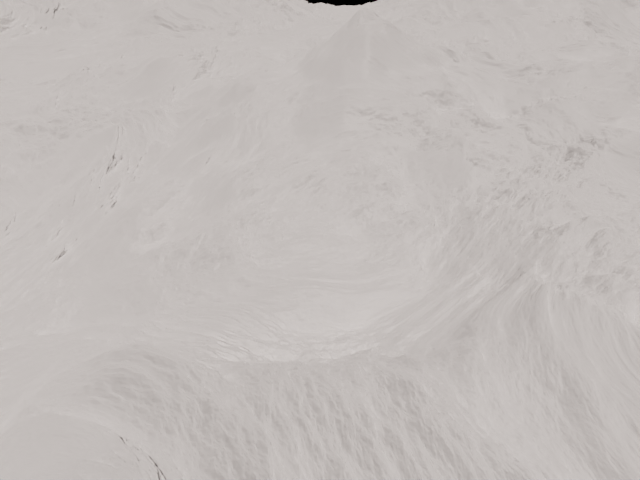}
\\
\includegraphics[width=0.15\textwidth]{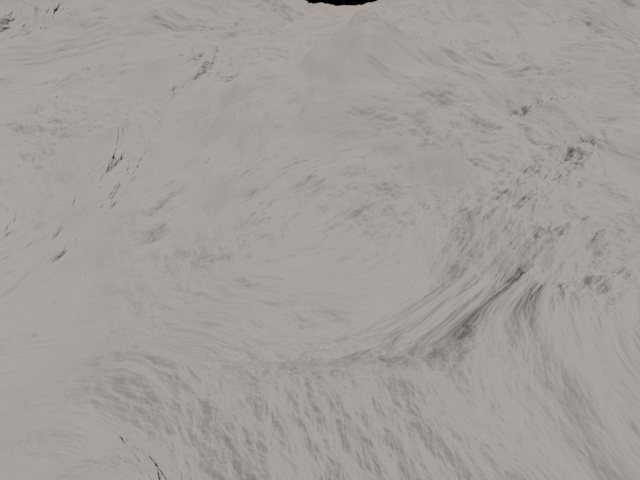}
\includegraphics[width=0.15\textwidth]{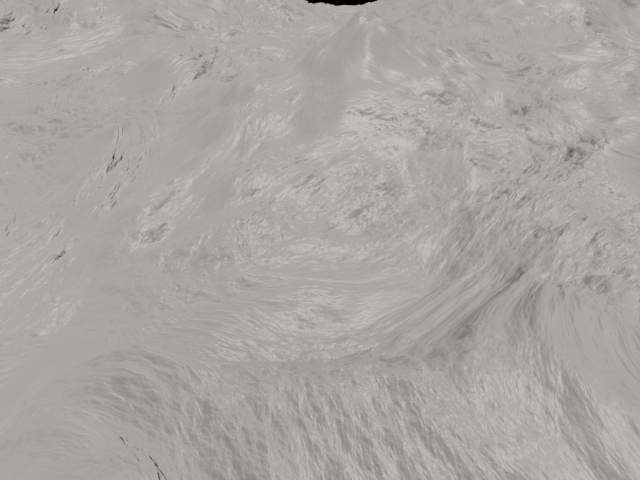}
\includegraphics[width=0.15\textwidth]{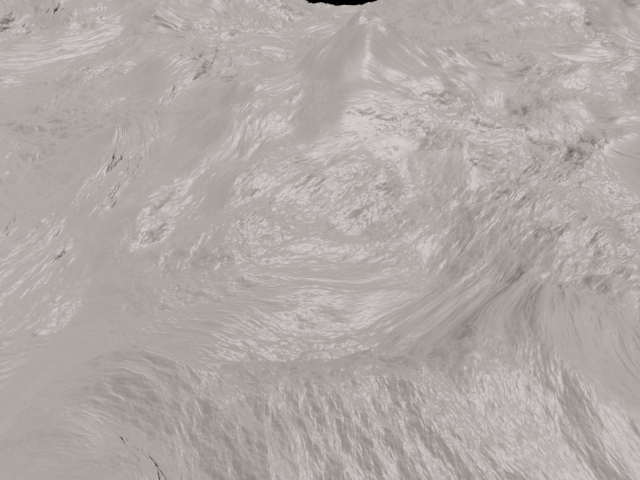}
\\
\includegraphics[width=0.15\textwidth]{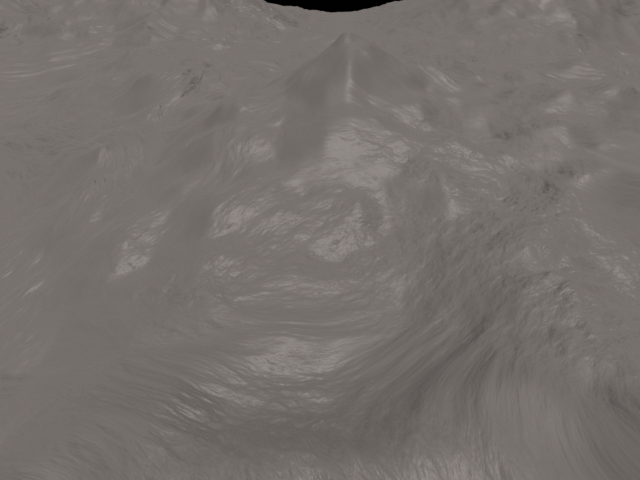}
\includegraphics[width=0.15\textwidth]{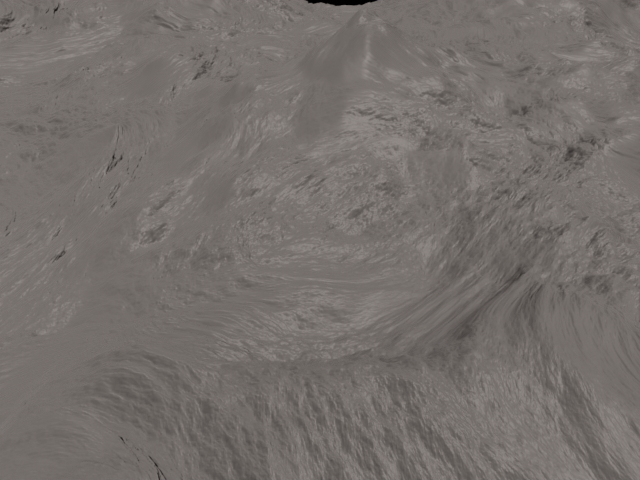}
\includegraphics[width=0.15\textwidth]{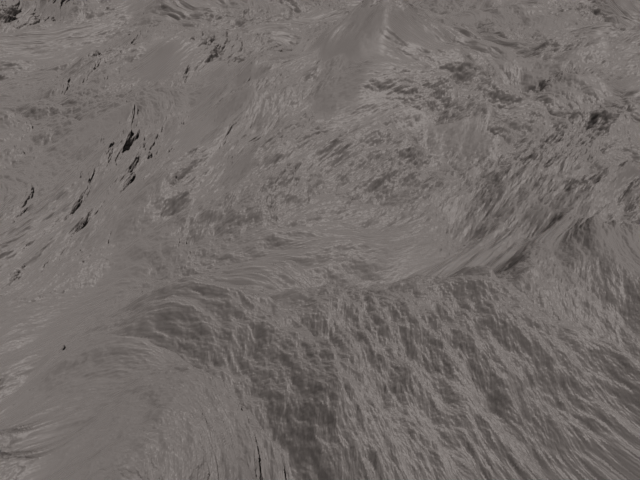}
\caption{Examples of generated texture with varying photometric properties. Top row: increasing albedo. Middle row: increasing specularity, Bottom row: increasing texture noise.}
\label{fig:examples_varying_photometric}
\end{figure}

Terrains depicting regolith of varying photometric properties are generated using procedural texture displacement as shown in Figures \ref{fig:examples_generative_subsurface}, \ref{fig:examples_varying_photometric}. 
Blender's Principled BSDF Shader is used to render the terrain-texture models in the suggested representations.


\ignore {
\begin{figure}
\centering
\includegraphics[width=3.25in]{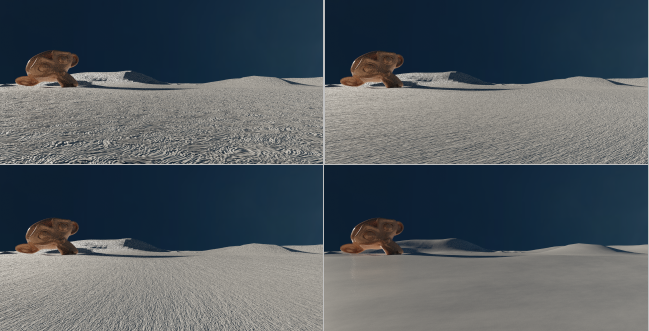}\\
\caption{Artistic illustration of procedural textures with varying levels of noise distortions.}
\label{Fig:roughnessScales}
\end{figure}

}


\ignore{
\subsection{Hapke BRDF}

Icy moon terrains exhibit strong surges in brightness at lower phase angles. Photometric studies (of Enceladus) indicate that the macroscopic roughness is significant at lower phase angles (angle between the Sun, target, and the lander). High geometric albedo (albedo at zero phase angle) indicates presence of less opaque material on the surface layer \cite{buratti1988enceladus}.  
These surface reflectance properties of the icy moon regolith needs to be addressed. Hapke BRDF \cite{hapke2012theory} is a parametric model considered a standard for modeling the backscattering and opposition surges from planetary regolith. 
The Hapke model incorporates texture properties of the regolith that include surface roughness, density, and porosity. Treating the regolithic composition as discrete particles, the Hapke model accounts for multiple scatterings within the regolithic layer. 
}


\subsection{Lighting}

GUISS provides the infrastructure to create physically correct lighting models that represent both direct and indirect light sources. Lighting computations are accurately performed using the Blender Cycles engine to produce a global illumination effect. Direct lighting dominates daytime and high-altitude observations, while indirect lighting becomes more prominent in observations taken closer to the surface. Light bouncing off Jupiter or Saturn also indirectly illuminate the respective icy moon surfaces. From a sampling mission perspective, studying the contribution of indirect lighting is equally important, where shadows with relatively low sharpness become significant.

In Blender, modeling the Sun involves specifying its color temperature, intensity of radiation in the visible spectrum (solar irradiance), and angular diameter, allowing for lighting the scene under conditions close to what is expected on icy worlds. The metadata on observation geometry, as viewed from specific points on the icy moons, is available through SPICE kernels provided by NASA's Navigation and Ancillary Information Facility (NAIF). To scale the planetary objects (indirect illuminators) down to the sampling workspace units, the observation geometries are modeled separately to represent accurate lighting conditions.

By placing the workspace at the origin of the virtual scene, the planetary objects are configured using the affine transformation between the body-frames of the icy world and its respective planet. The metadata on relative distances and sizes are utilized to represent the apparent sizes of the planetary bodies as computed in Eq. \ref{eq:apparentSizes}. 


\begin{equation} \label{eq:apparentSizes}
\frac{\text{Apparent Size (deg)}}{57.3} = \frac{\text{Diameter of the object}}{\text{Distance from the object}}
\end{equation}

\ignore{
\begin{figure}
\centering
\includegraphics[width=3.25in]{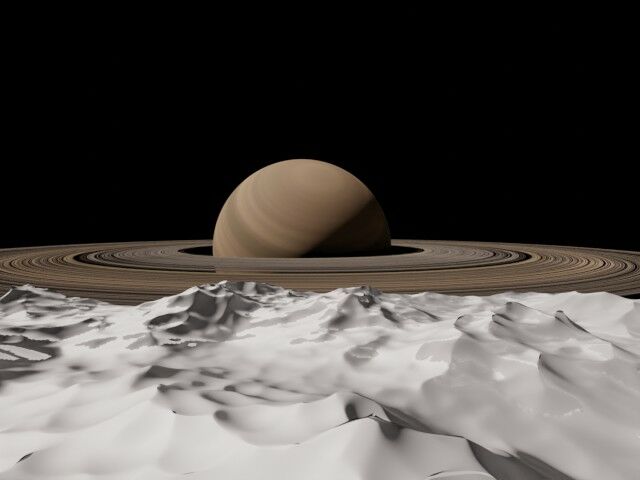}\\
\caption{{Scaled planetary representation of Enceladus terrain with Saturn as a source of indirect illumination.}}
\label{Fig:illuminationModel}
\end{figure}
}

\begin{figure}
     \centering
     \begin{subfigure}[b]{0.23\textwidth}
         \centering
         \includegraphics[width=\textwidth]{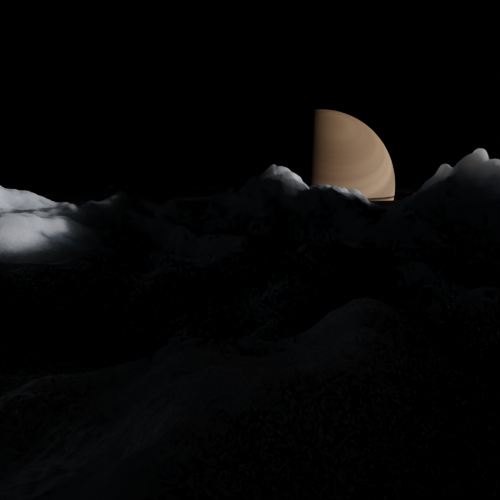}
         \caption{Elevation: $0\deg$.}
         \label{fig:elevation0}
     \end{subfigure}
     \hspace{2mm}
     \begin{subfigure}[b]{0.23\textwidth}
         \centering
         \includegraphics[width=\textwidth]{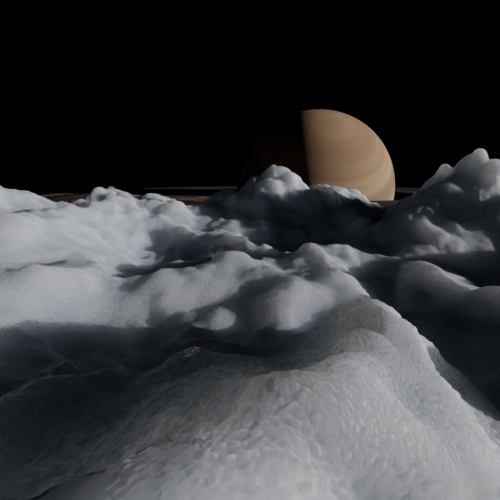}
         \caption{Elevation: $10\deg$.}
         \label{fig:elevation10}
     \end{subfigure}
     \hfill
     \begin{subfigure}[b]{0.23\textwidth}
         \centering
         \includegraphics[width=\textwidth]{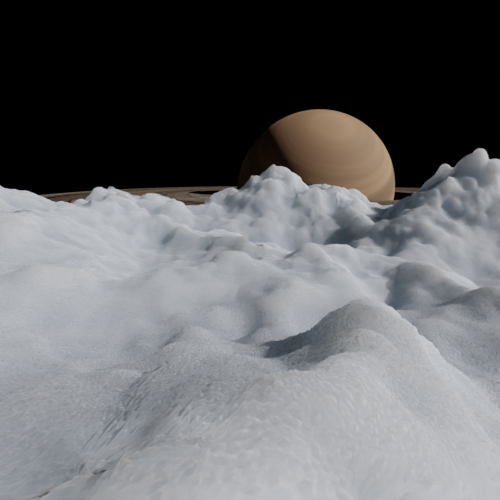}
         \caption{Elevation: $35\deg$.}
         \label{fig:elevation35}
     \end{subfigure}
     \hspace{2mm}
     \begin{subfigure}[b]{0.23\textwidth}
         \centering
         \includegraphics[width=\textwidth]{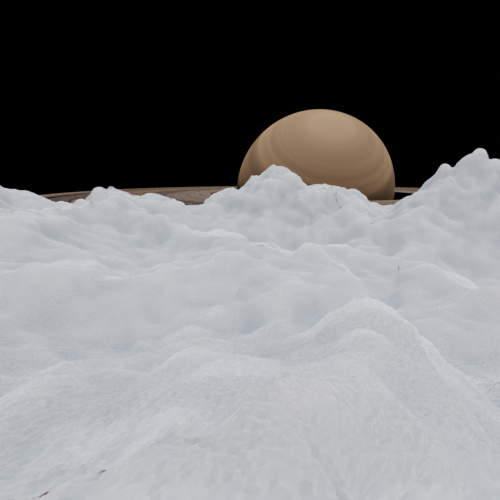}
         \caption{Elevation: $90\deg$.}
         \label{fig:elevation90}
     \end{subfigure}
        \caption{Surface illumination at varying Sun elevation angles.}
        \label{Fig:elevationAngles}
\end{figure}

Figure \ref{Fig:elevationAngles} illustrates the icy moon surface at varying angles of Sun elevation. Lighting from planetary reflections and subsurface scattering influences terrain illumination even when no direct light is incident upon the surface.



\subsection{Camera Model}

The framework can generate high-fidelity simulations and metadata for visual perception using Blender Cycles. Blender supports modeling camera intrinsic elements such as sensor size, focal length, and lens distortion. Additional capabilities include depth of field for blurring effects and shutter speed for simulating motion blur. Various lens models such as perspective, orthographic, and panoramic lens models are also supported.
Blender natively supports stereo rendering with two virtual cameras with a configurable baseline distance. It renders depth images using a z-buffer, where the z-coordinate values of each pixel are stored during rendering. This z-buffer information is then utilized to generate accurate depth maps, serving as ground truth for applications such as depth sensing and 3D reconstruction. 
Figure \ref{fig:qualitative_examples} (second column) demonstrates the ground truth depth synthesized from the sensor model.

The shortcomings include capturing details in high-contrast environments, such as the bright surface of icy world against a dark background. 
While Blender simulates exposure settings for a scene, it does not represent physical exposure time. Settings in a planetary environment with an extensive dynamic range require careful and manual adjustments for a realistic output.
Blender does simulate noise in the rendered images, but this simulation is not specific to noise models of real representative sensors used in planetary missions. Achieving accurate simulation often requires additional scripting in Blender, contributions from the community in the form of add-ons, as well as filtering and post-processing to align with specific sensor models.

%% file: Sections/05_stereo.tex
\section{Stereo Depth Estimation}


Our main motivation for the development of GUISS is to assess the performance of a candidate perception system that can facilitate autonomous sampling. Such a perception system would be required to capture the visual appearance and retrieve the geometry of the terrain within the workspace of the lander. These capabilities are important both for collision avoidance with the environment and identifying interesting sites for excavation during sampling operations. 

In this work, we instantiate our perception system as a simulated stereo camera provided by Blender following the pinhole camera model. The cameras have a sensor width of $60$mm, focal length of $32$mm, and a baseline distance of $0.25$m. We prioritize the investigation of a stereo-based visual perception system over other candidates such as active sensing (e.g., LiDAR), as it has been successfully used in prior missions (e.g., MER \cite{goldberg2002stereo}, Mars2020), and it has been identified by prior work on Europa Lander \cite{bowkett2021functional,ono2020ariel} as the primary vision system.

While stereo vision has been successful on other planetary bodies such as Mars where the terrain is texture-rich, icy worlds such as Europa and Enceladus are expected to have visually challenging environments with highly reflective materials, feature-deprived regions, or highly textured surfaces with repetitive patterns. A very high albedo ($\geq0.8$) is also expected on Enceladus which can blind the camera from any terrain features. Traditional stereo matching algorithms \cite{scharstein2002taxonomy} typically employed in missions tend to struggle under these conditions as they rely on matching pixels (or local areas) between the images.

In response to these issues, alternative stereo-based depth estimation methods \cite{laga2020survey} have been proposed using deep learning techniques. The main advantage of these methods is that they are able to leverage prior knowledge over the 3D structure of the world by consuming large amounts of data. This leads to the derivation of highly complex visual representations enabling them to identify visual patterns from prior experience. 

\subsection{Baselines}
We are interested in evaluating the performance of both classical stereo matching methods and contemporary deep learning-based approaches. Note that for the latter we are using the provided off-the-shelf models which were trained on typical in-the-wild datasets such as KITTI \cite{geiger2012we,menze2015object} and SceneFlow \cite{mayer2016large} and have not seen any icy moon related data during training. We use the following baselines:
\begin{itemize}
    \item \textbf{StereoBM \cite{scharstein2002taxonomy,kanade1995development}.} A classical method that performs Block Matching (BM), i.e., it uses Sum of Absolute Differences (SAD) to find matches between local windows from two rectified images along a certain scan line. There is a number of parameters that can affect StereoBM's performance, with the two most important parameters being the number of disparities to be searched and the size of the sliding window. The values of these parameters are empirically determined as 96 and 49 respectively. In our experiments, we use the OpenCV implementation of StereoBM.
    \item \textbf{JPLV \cite{goldberg2002stereo}.} This is the JPL-developed stereo depth estimation method that has been successfully used in prior missions such as the Mars Exploration Rover (MER). It optimizes classical methods by using a scale pyramid and computing the Laplacian of each image to remove any pixel intensity bias.
    \item \textbf{DSMNet \cite{zhang2020domain}.} Domain-invariant Stereo Matching Networks (DSMNet) is a deep learning-based method that was developed with the objective of improved generalization to new visual domains. To achieve this, the method introduced a normalization layer that regulates the distribution of features between domains, and a learnable graph-based layer that extracts structural information from the image.
    \item \textbf{IGEV \cite{xu2023iterative}.} Iterative Geometry Encoding Volume for Stereo Matching (IGEV) is a state-of-the-art deep learning-based method, achieving top performance on current publicly available benchmarks \cite{aanaes2016large,geiger2012we,menze2015object}. This method employs 3D convolutions to produce an initial disparity map, which in turn is refined through an iterative optimization process using Convolutional Gated Recurrent Units (ConvGRUs).
\end{itemize}

\subsection{Datasets}
Following the capabilities of our icy moon surface simulation as described in the Section~\ref{sec:rendering_capabilities}, we generate several datasets depicting different visual hypotheses of Europa and Enceladus. Given that our current understanding of the terrain specifications of Europa and Enceladus at the meter and sub-meter scale is somewhat limited (see Section~\ref{sec:terrain_design_reqs}), we need to cover a wide spectrum of visual hypotheses, especially the more challenging cases for stereo matching (smooth texture-less snow/ice surfaces with low illumination and high albedo). To this end, we evaluate stereo-based depth estimation under varying scene parameters such as terrain geometry, textures, lighting, and photometric properties of the surface. In order to ground both our simulated data and experimental results, we additionally render and evaluate on imagery of scenes reconstructed from real images. Our evaluation data is described as follows:
\begin{itemize}
    \item \textbf{Procedural terrain variation.} We generate several hypotheses with respect to terrain geometry which span smooth plane-like surfaces to highly rugged terrain resembling penitentes (see Figures~\ref{Fig:terrain_procedural} and ~\ref{Fig:Gaea_procedural}). 
    To enrich the terrain we also procedurally add rock formations of various sizes and density (see Figure~\ref{fig:examples_rocks}).
    \item \textbf{Texture variation.} We aim to capture imagery with a vast variety of textures that correspond to snow and icy surfaces. This is achieved in two ways: 1) We use publicly available texture images and albedo maps \cite{quixel} (see Figure~\ref{fig:examples_texture_variation}), and 2) we procedurally generate textures by varying noise parameters (see Figure~\ref{fig:examples_varying_photometric} bottom row). 
    \item \textbf{Photometric properties.} Using Blender's Principled BSDF we vary the following parameters: albedo (base color), subsurface factor, specularity, and transmission (see Figures~\ref{fig:examples_generative_subsurface} and ~\ref{fig:examples_varying_photometric}).
    \item \textbf{Lighting variation.} Lighting is varied in two ways: 1) Sunlight energy spans $4.140 \text{W/m}^2$ (Enceladus) to $50.26 \text{W/m}^2$ (Europa), and 2) we use different sunlight elevation and azimuth angles ($0^\circ$, $30^\circ$, $60^\circ$). The angular diameter of the sunlight is constant at $0.01$ radians. See Figure~\ref{Fig:elevationAngles} for examples.
    \item \textbf{Scene Reconstructions.} Finally, we generate imagery from real reconstructed scenes of the Matanuska and Athabasca glaciers in Alaska and Alberta, respectively. The images were collected from a Parrot Anafi drone and used PIX4D photogrammetry software \cite{pix4d} for reconstruction. The purpose of these scenes is to provide Earth analogues to Europa and Enceladus and cover real visual representations of various snow and ice surfaces. Figure~\ref{fig:examples_scene_reconstructions} shows several examples.
\end{itemize}
In total we generated 5020 stereo images for our stereo depth estimation evaluation along with their corresponding ground-truth depth maps. 3060 of these images are rendered from the real reconstructed scenes, while the remaining 1960 are rendered following the aforementioned varying scene parameters.

\subsection{Experimental Evaluation}

Four metrics are used to evaluate the quality of the depth estimation from our baselines. In the following definitions $y_i$ is the ground-truth depth and $\hat{y}_i$ is the predicted depth of pixel $i$. $N$ denotes the total number of test examples (in pixels). 
\begin{itemize}
    \item \textbf{L1 Error.} Absolute difference between ground truth and predicted depth:
    \begin{equation}
        L_1(\hat{y}, y) = \frac{1}{N} \sum^N_{i=1} |\hat{y}_i-y_i|
    \end{equation}

    \item \textbf{L1 error rate 10.} Percentage of pixels with $>10$ cm L1 error.
    
    \item \textbf{Scale-invariant Root Mean Squared Error (si-RMSE).} Error metrics such as L1 are affected by the global scale of a scene. Following \cite{eigen2014depth,li2018megadepth} who observed that the ratios of pairs of depths are preserved under scaling we adopt the si-RMSE which computes the MSE of the difference between all pairs of log-depths:
    \begin{equation}
        D(\hat{y}, y) = \frac{1}{N} \sum^N_{i=1} (d_i)^2 - \frac{1}{N^2} (\sum^N_{i=1} d_i)^2
    \end{equation}
    where $d_i=\log\hat{y}_i - \log y_i$.

    \item \textbf{Depth Ordering Disagreement (DOD).} Besides measuring absolute error, it is also important to evaluate how well a baseline predicts the structure of the scene. This is captured by measuring the disagreement rate between predicted depth $\hat{y}$ ordinal relations and ground-truth depth $y$ ordinal relations. Following \cite{chen2016single,li2018megadepth}, this is defined as:
    \begin{equation}
    {DOD}(\hat{y},y)=\frac{1}{N}\sum_{i,j\in P} {\mathds{1}} ({ord}(\hat{y}_i, \hat{y}_j ) \neq ord(y_i, y_j)),
    \end{equation}
    where $P$ is the set of pixel index pairs and $ord(\cdot,\cdot)$ is one of three depth ordering relations (further-than, closer-than, and same-depth-as):
    \begin{equation}
    ord(y_i,y_j) =
    \begin{cases}
      1 & \text{if $\frac{y_i}{y_j} > 1 + \tau$,}\\
      -1 & \text{if $\frac{y_i}{y_j} < 1 - \tau$,}\\
      0 & \text{otherwise}, 
    \end{cases}
    \end{equation}
    with tolerance $\tau=0.01$ m.
    
\end{itemize}

\begin{figure*}[t]
\centering
\includegraphics[width=0.99\textwidth]{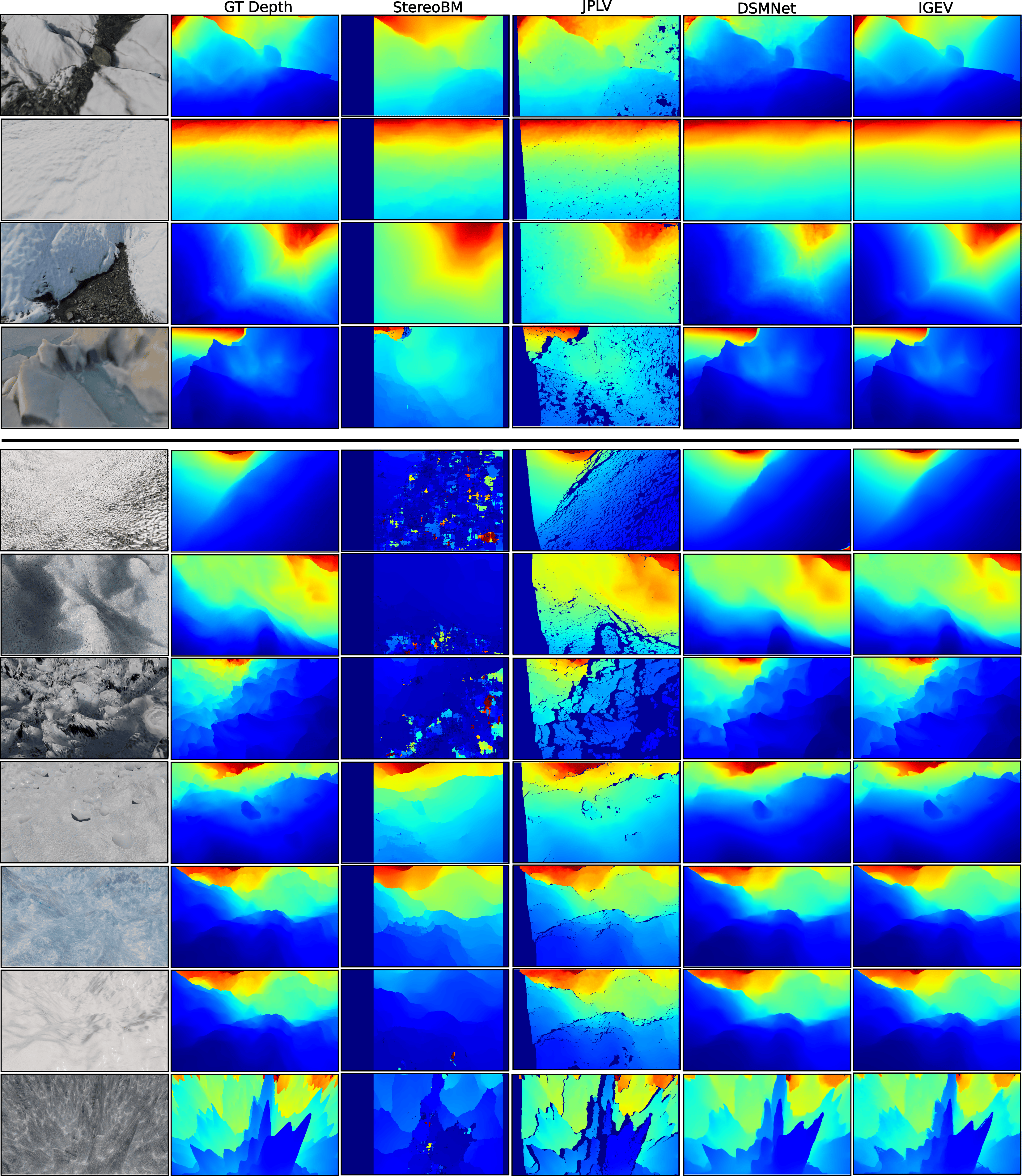}
\caption{Qualitative stereo depth estimations from real reconstructed (top) and synthetic (bottom) scenes.}
\label{fig:qualitative_examples}
\end{figure*}

\begin{table*}[t]
    \centering
    \scalebox{1.0}{
    \begin{tabular}{lcccc}
        \hline
        \textbf{All} & L1 (m) $\downarrow$ & L1 rate 10 (\%) $\downarrow$ & si-RMSE $\downarrow$ & DOD (\%) $\downarrow$ \\
        \hline
        StereoBM \cite{scharstein2002taxonomy} & 1.13 & 72.5 & 0.30 & 12.4 \\
        JPLV \cite{goldberg2002stereo} & 0.55 & 73.9 & 0.74 & 13.1 \\
        DSMNet \cite{zhang2020domain} & 0.29 & 66.1 & 0.03 & 2.6 \\
        IGEV \cite{xu2023iterative} & 0.26 & 64.4 & 0.03 & 2.3 \\
        \hline
        \hline
        \textbf{Reconstructed Scenes} & & & & \\
        \hline
        StereoBM \cite{scharstein2002taxonomy} & 0.73 & 69.2 & 0.11 & 5.7 \\
        JPLV \cite{goldberg2002stereo} & 0.58 & 74.5 & 0.67 & 7.2 \\
        DSMNet \cite{zhang2020domain} & 0.39 & 71.7 & 0.02 & 2.3 \\
        IGEV \cite{xu2023iterative} & 0.34 & 70.2 & 0.02 & 1.9 \\
        \hline
        \hline
        \textbf{Synthetic Scenes} & & & & \\
        \hline
        StereoBM \cite{scharstein2002taxonomy} & 1.76 & 77.6 & 0.6 & 22.9 \\
        JPLV \cite{goldberg2002stereo} & 0.48 & 73.1 & 0.86 & 22.3 \\
        DSMNet \cite{zhang2020domain} & 0.14 & 57.5 & 0.04 & 3.1 \\
        IGEV \cite{xu2023iterative} & 0.14 & 55.4 & 0.04 & 3.1 \\
        \hline
    \end{tabular}
    }
    \caption{Performance evaluation of our baselines over the entirety of our evaluation data (top), the real reconstructed scenes (middle), and the synthetic scenes (bottom).}
    \label{tab:results_all}
\end{table*}

\subsection{Results}
Our evaluation of the baselines over our generated data (both synthetic and from real scenes) is shown in Table~\ref{tab:results_all} (top), while qualitative examples are illustrated in Figure~\ref{fig:qualitative_examples}. It is immediately apparent that the deep learning-based baselines outperform the classical method of StereoBM by a large margin on all metrics, while JPLV is comparable on L1 error in the real reconstructed scenes. It is important to note that DSMNet and IGEV achieve this performance even though they were trained on visually dissimilar domains \cite{geiger2012we,menze2015object,mayer2016large} to the evaluation data. Specifically, there is an $87$ cm and $10.1\%$ gap from StereoBM in L1 Error and DOD respectively. This can be interpreted as the StereoBM having both poorer performance when it comes to absolute depth estimation and with respect to the structure of the scene. 
Comparing StereoBM to JPLV we notice that the latter achieves smaller errors in L1, while it underperforms on si-RMSE. This suggests that JPLV is more accurate at longer distances but it makes consistent errors at multiple scales. 
Furthermore, the $2.3\%$ DOD result of IGEV suggests that it can predict the right depth ordering of the pixels in the scenes almost perfectly. This is undoubtedly a result of leveraging prior knowledge over scene structure that is distilled during training on real in-the-wild data.

While the L1 Error performance of DSMNet and IGEV is much better than StereoBM and JPLV, the $0.29$ and $0.26$ error respectively is still not ideal for excavation purposes. However, this high value can be explained by the fact that most reconstructed scenes are larger in scale than the expected workspace of the lander, hence contributing to higher L1 errors. This is where si-RMSE is useful because it treats all scalar multiples of a pixel depth to have the same error. DSMNet and IGEV achieve a very low $0.03$ si-RMSE, suggesting that if the scene was constrained to a smaller workspace, the L1 Error would be lower as well.

We investigate this further by showing results separately on the real reconstructed scenes (Table~\ref{tab:results_all} middle) and the synthetic data (Table~\ref{tab:results_all} bottom). There is a clear discrepancy between these results, with StereoBM achieving significantly lower error on the reconstructed scenes than what it achieves on the synthetic data, while the deep learning-based methods follow the opposite route. This is related to the presence of more texture-rich materials in the real scenes and it highlights the difficulty of our synthetic scenes where StereoBM is frequently failing (see examples in Figure~\ref{fig:qualitative_examples}).

\begin{figure*}[t]
\centering
\includegraphics[width=0.3\textwidth]{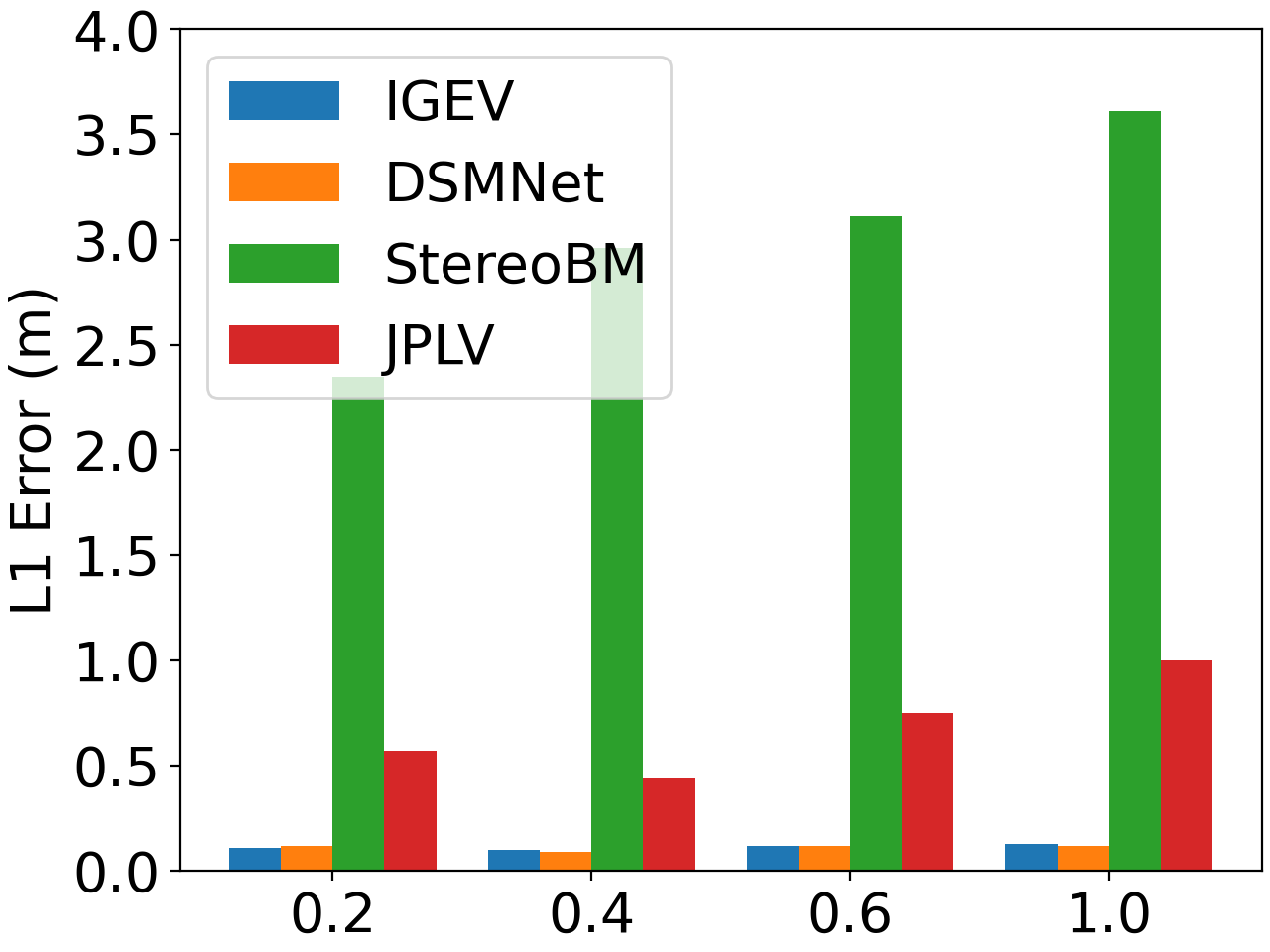}
\includegraphics[width=0.3\textwidth]{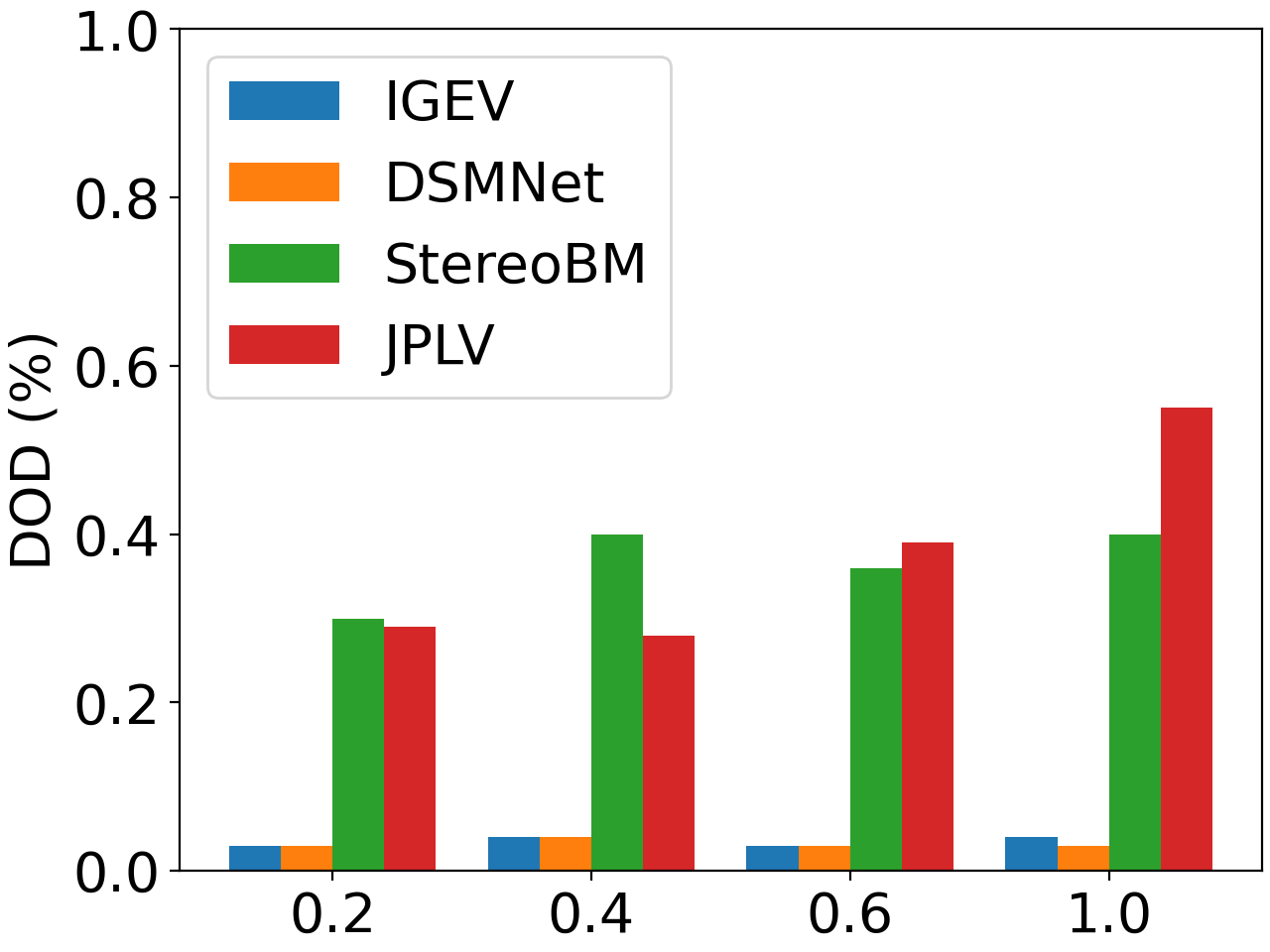}
\includegraphics[width=0.3\textwidth]{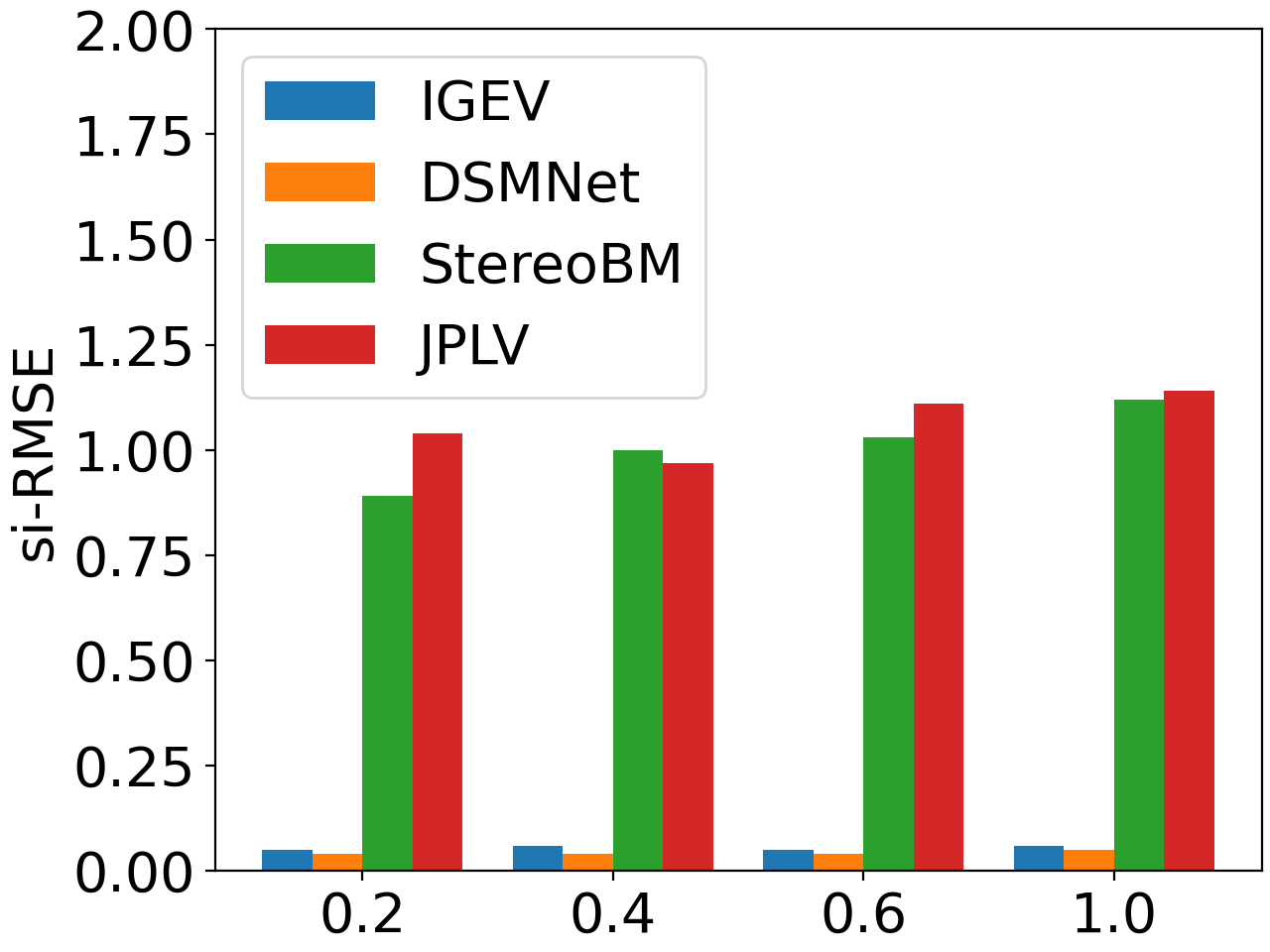}
\caption{Results with varying values of albedo.}
\label{fig:res_albedo}
\end{figure*}

Additionally, 
we evaluate the baselines under varying values of albedo (base color) in our synthetic scenes, with results shown in  
Figure~\ref{fig:res_albedo}.
We notice that StereoBM and JPLV performance degrades with increasing values of albedo as higher uniform intensities reduce the visible texture and making it harder for the block matching to find correspondences. On the other hand, the deep learning-based methods demonstrate robustness to high albedos which are expected on Europa and Enceladus.

Finally, we argue that the performance of the depth estimation can be enhanced in two ways: 1) By finetuning the deep learning-based methods on our data, and 2) by incorporating sparse measurements from an active sensor (e.g., LiDAR) that can be used to refine the dense prediction of our baselines. Both of these approaches are outside the scope of this work.

\textbf{Computational Complexity of Rendering.}
Our simulation requires around 3 seconds to render a stereo image (and corresponding depth image) using the real reconstructed scenes on a mid-range laptop with an Intel i7 CPU. For the synthetic scenes on the same system we require around 1 minute due to the additional computational burden of using the Principled BSDF, and the procedural terrain and textures.

\textbf{Computational Complexity of Stereo Depth Estimation.} We show the runtimes of the stereo estimation baselines in Table~\ref{tab:runtimes}. These numbers were captured on a mid-range laptop with an Intel i7 CPU. For the deep-learning methods we used an NVIDIA RTX 3080 laptop GPU.

\begin{table}[t]
    \centering
    \scalebox{0.78}{
    \begin{tabular}{l|c|c|c|c}
        \hline
         & StereoBM \cite{scharstein2002taxonomy} & JPLV \cite{goldberg2002stereo} & DSMNet \cite{zhang2020domain} & IGEV \cite{xu2023iterative} \\
        \hline
        Runtime (s) & 0.02 & 0.15 & 0.45 & 0.34 \\
        \hline
    \end{tabular}
    }
    \caption{Runtimes of the stereo depth estimation baselines used in our experiments. The numbers shown are for the estimation over a stereo pair with $640\times480$ images.}
    \label{tab:runtimes}
\end{table}

%% file: Sections/06_Conclusions.tex
\section{Conclusions}
   
This paper outlines the imperative need for a rendering utility capable of realistically simulating the challenging environments found on icy moons. In response, we introduce the Graphical Utility for Icy Moon Surface Simulations (GUISS), designed for photorealistic stereo vision simulations towards evaluation of sampling perception algorithms. Leveraging Blender's robust capabilities, GUISS facilitates the modeling of terrains and textures for icy moon surfaces through procedural methods. The utility includes models that incorporate realistic lighting scenarios and stereo imaging capabilities, coupled with ground truthing features embedded within the software. GUISS is designed with the aim to create versatile datasets that span the spectrum of potential observations on icy worlds. Ongoing efforts are dedicated to addressing the persistent challenges in achieving high-fidelity rendering to meet specific design requirements. The work emphasizes the importance of performance benchmarking for perception systems, particularly in-depth estimation, within high dynamic range environments like those encountered on icy moons.

Moreover, the paper explores the evaluation of state-of-the-art depth estimation algorithms using simulated imagery. Our experimental results demonstrate the superiority of the deep learning-based methods in these challenging conditions and pose an argument to support the usage of machine learning and AI in future missions to icy worlds.

%% file: Sections/07_FutureWork.tex
\section{Limitations and Future Work}


GUISS employs path tracing for high-fidelity icy landscape rendering, but it is non-real-time.
Currently in active development, there are certain aspects of the framework where it currently falls short:
\begin{itemize}

    \item \textbf{Validation challenges.} 
   The lack of surface imagery poses challenges for validating GUISS simulations. Ongoing efforts focus on improving validation through large-distance simulations and other planetary observations.
   
    \item \textbf{Photometric studies.} While GUISS can generate large datasets with versatility in terrains and textures, realistic photometric studies are not completely evaluated. Efforts are ongoing to enhance the photometric modeling, exploring relevant Bidirectional Reflectance Distribution Functions (BRDFs) such as the Hapke model.
    
    \item \textbf{SPICE Kernels.} GUISS currently does not fully support the integration of SPICE toolkit-specific position and lighting angle observations. 
    
    \item \textbf{Active sensing.} GUISS is exploring the incorporation of realistic active sensor models, such as LiDAR or structured light, to complement the stereo system. While LiDAR is already part of the simulations (not discussed in this paper), a realistic sensor model is yet to be fully integrated.

\end{itemize}

These ongoing efforts reflect a commitment to addressing the current limitations and further enhancing the capabilities of GUISS, along with corresponding perception sensor analysis for icy worlds.




%% file: Sections/08_Acknowledgements.tex
\section{Acknowledgement}

{\textcopyright}2023 California Institute of Technology. The research described in this paper was carried out at the Jet Propulsion Laboratory (JPL), California Institute of Technology, under a contract with the National Aeronautics and Space Administration. Government sponsorship acknowledged. 

Texas A\&M University is duly acknowledged for providing support towards visiting research program at the JPL.

JPL technologists Issa Nesnas, Yang Cheng, Asher Elmquist, Travis Driver, Gregory Griffin, Spencer Diehl, Ishan Mishra, Ashish Goel, Anup Katake, Reg Willson, Micheal Swan, Deegan Atha, Daniel Moreno, Samuel Howell are acknowledged for insightful discussions on the development.

%% file: Sections/biography.tex
\thebiography

\begin{biographywithpic}
{Ramchander Bhaskara}{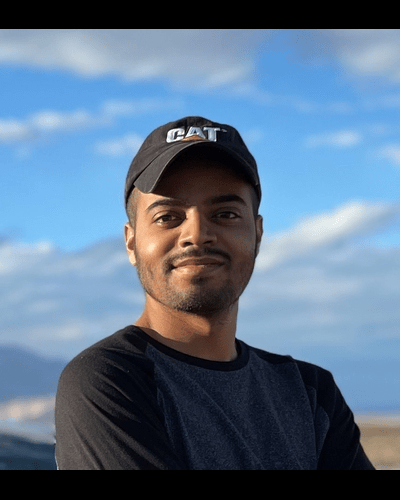} 
is a PhD student in Aerospace Engineering at Texas A\&M University. His dissertation is on real-time signal processing and state estimation for high-rate sensing using FPGAs. His research interests also include computer graphics and vision for space scene visualization and spacecraft navigation. 
Ram interned at JPL during Summer 2023.
\end{biographywithpic} 

\begin{biographywithpic}
{Georgios Georgakis}{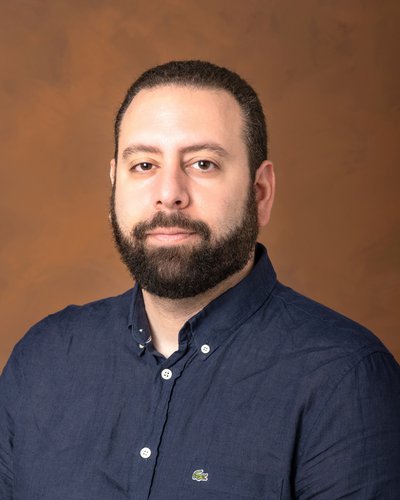}
is a Robotics Technologist at NASA's Jet Propulsion Laboratory, California Institute of Technology. Before joining JPL he was a Postdoctoral Researcher and Part-time Lecturer in the GRASP lab at University of Pennsylvania. He earned his PhD in Computer Science from George Mason University in 2020. His research interests lie at the intersection of computer vision and machine learning that have applications in robotics.
\end{biographywithpic}

\begin{biographywithpic}
{Jeremy Nash}{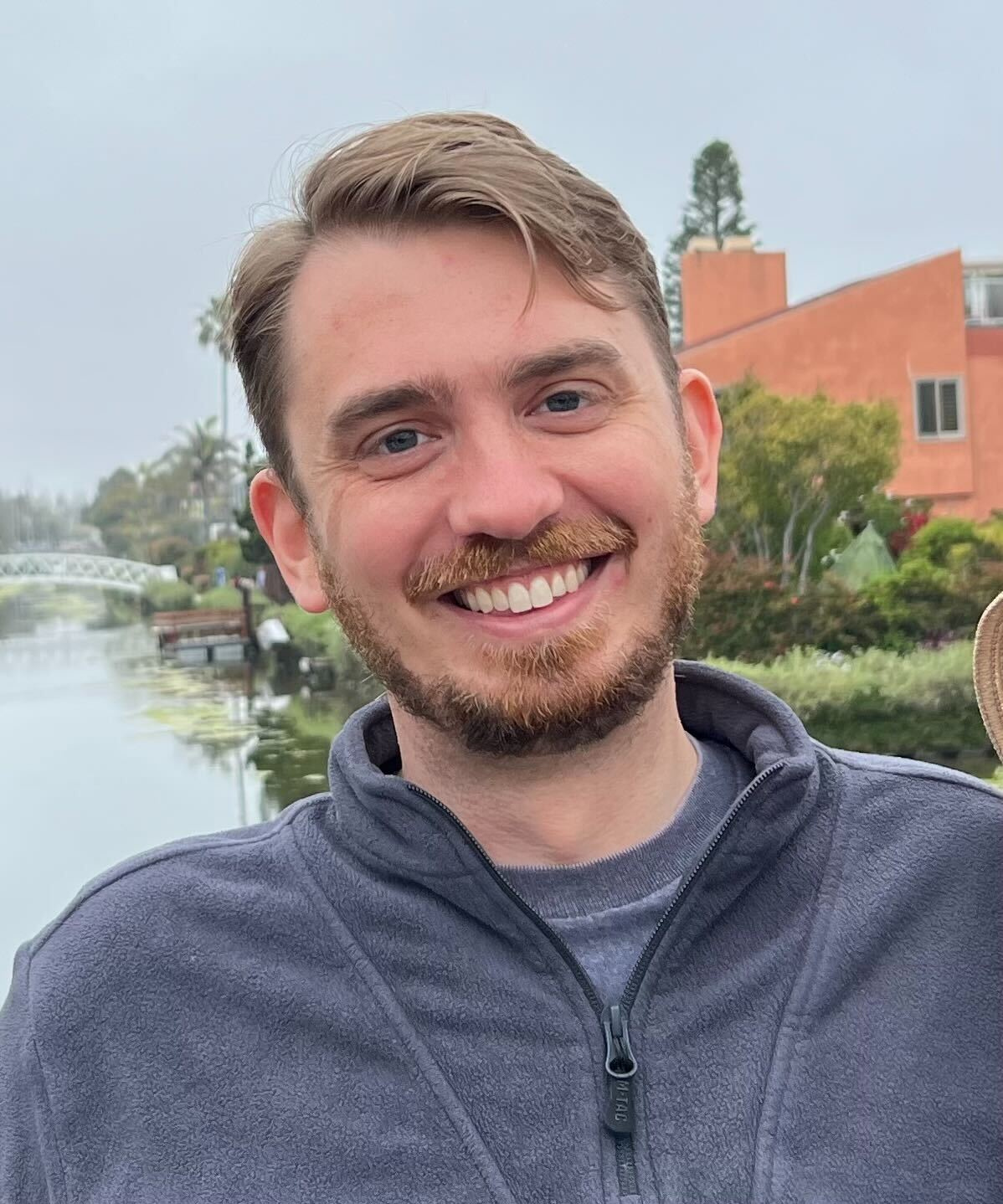} is a Group Lead of the Aerial and Orbital Image Analysis group at the Jet Propulsion Laboratory. His recent work includes creating Mars maps for the Sample Retrieval Lander (SRL) Entry, Descent, and Landing (EDL) system, and global localization for the Perseverance rover. 
\end{biographywithpic}

\begin{biographywithpic}
{Marissa Cameron}{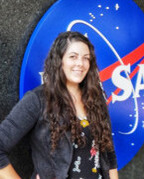} received her PhD in Geology and Geophysics and has in-depth knowledge of icy satellite terrains, experience mapping planetary surfaces, and familiarity with a wide range of software in data processing, statistical analysis, and visualization. She is currently working on terrain specification to enable landing and sampling site selection in support of the Europa Lander pre project.
\end{biographywithpic}

\begin{biographywithpic}
{Joseph Bowkett}{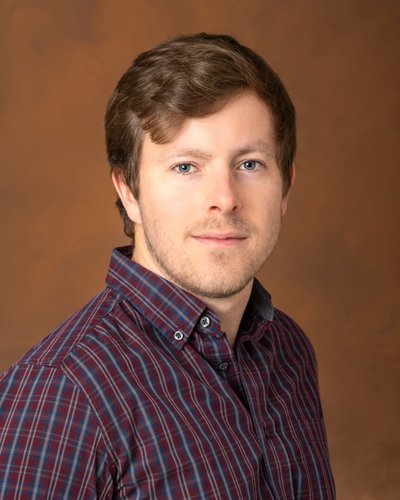} is a Robotics Technologist and Group Lead of the Robotic Manipulation and Sampling group at NASA's Jet Propulsion Laboratory, California Institute of Technology. He completed a PhD in the Burdick group of the Mechanical \& Civil Engineering Department at the California Institute of Technology. His interests focus around what is termed behavior level or ‘functional’ autonomy for robotic tasks, particularly in regard to grasping and manipulation, leveraging both proprioception and exteroception to build understanding of unstructured task spaces.
\end{biographywithpic}

\begin{biographywithpic}
{Adnan Ansar}{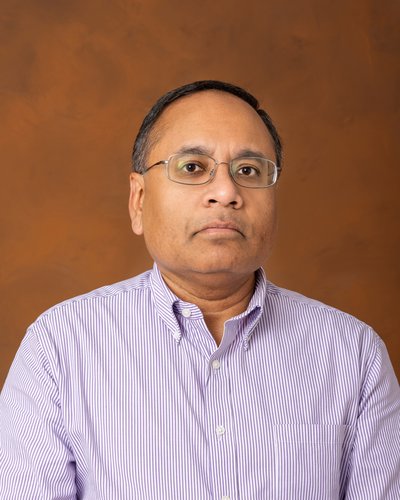} received his PhD in Computer Science from the GRASP Laboratory at the University of Pennsylvania in 2001. He has been a member of the Mobility and Robotic Systems Section at NASA’s Jet Propulsion Laboratory (JPL) since 2002 and currently manages the Aerial and Orbital Image Analysis Group within that Section. While at JPL, his research has included image-based position estimation, camera calibration, stereo vision, structure from motion, multi-modal data registration, and orbital mapping for terrain relative navigation.
\end{biographywithpic}

\begin{biographywithpic}
{Manoranjan Majji}{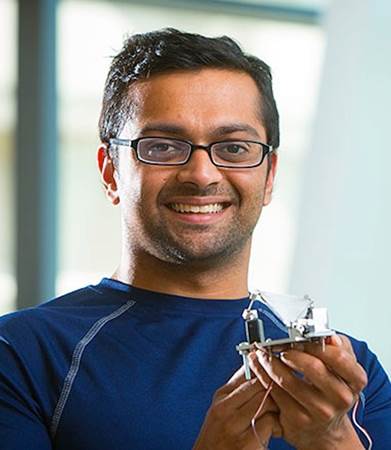} earned his Ph. D. in Aerospace Engineering from Texas A\&M University in 2009. He is an Associate Professor of Aerospace Engineering Department at Texas A\&M University and directs the Land, Air and Space Robotics (LASR) laboratory. He teaches graduate and undergraduate courses in mechanics, control, systems analysis, estimation of dynamical systems and astrodynamics. He received the Milton Plesur award for excellence in teaching from the University at Buffalo, State University of New York. He is a Fellow of the American Astronautical Society (AAS).
\end{biographywithpic}

\begin{biographywithpic}
{Paul Backes, Ph.D.}{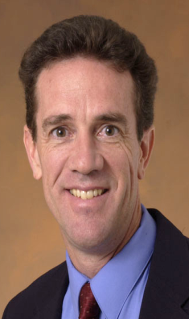}  is Group Supervisor of the Robotic Manipulation and Sampling group at Jet Propulsion Laboratory, where he has been since 1987.  He received the BSME degree at U.C. Berkeley in 1982 and Ph.D. in ME from Purdue University in 1987. His awards include NASA Exceptional Engineering Achievement Medal (1993), JPL Award for Excellence (1998), NASA Software of the Year Award (2004), IEEE Robotics and Automation Technical Field Award (2008), and NASA Exceptional Service Award (2014).
\end{biographywithpic}